\documentclass{article}

    \usepackage[preprint]{neurips_2023}

\usepackage[utf8]{inputenc} 
\usepackage[T1]{fontenc}    
\usepackage{hyperref}       
\usepackage{url}            
\usepackage{booktabs}       
\usepackage{amsfonts}       
\usepackage{nicefrac}       
\usepackage{microtype}      
\usepackage{xcolor}         
\usepackage{natbib}

\usepackage{amsmath}
\usepackage{amsthm}
\usepackage{graphicx}
\usepackage{tikz}
\DeclareMathOperator*{\argmin}{arg\,min}
\DeclareMathOperator*{\argmax}{arg\,max}
\DeclareMathOperator*{\diam}{diam}
\DeclareMathOperator*{\Vol}{Vol}
\DeclareMathOperator*{\ddim}{ddim}
\DeclareMathOperator*{\var}{var}

\newtheorem{lemma}{Lemma}
\newtheorem{definition}{Definition}
\newtheorem{theorem}{Theorem}
\newtheorem{corollary}{Corollary}

\usepackage[toc,page]{appendix}

\usepackage{booktabs}

\begin{document}




\title{Fr\'echet random forests for metric space valued regression\\ with non euclidean predictors}

\author{Louis Capitaine\\
       epoch intelligence\\
       Univ. Bordeaux, INSERM, INRIA, BPH, U1219\\
       F-33000 Bordeaux, France\\
       \texttt{louis.capitaine@epoch-intelligence.fr}
       \And
       J\'er\'emie Bigot\\
       Univ. Bordeaux, CNRS, Bordeaux INP, IMB, UMR 5251\\
       F-33400 Talence, France\\
       \texttt{jeremie.bigot@math.u-bordeaux.fr}
       \And 
       Rodolphe Thi\'ebaut\\
       Univ. Bordeaux, INSERM, INRIA, BPH, U1219\\
       F-33000 Bordeaux, France\\
       \texttt{rodolphe.thiebaut@u-bordeaux.fr}
       \And
       Robin Genuer\\
       Univ. Bordeaux, INSERM, INRIA, BPH, U1219\\
       F-33000 Bordeaux, France\\
       \texttt{robin.genuer@u-bordeaux.fr}
       }


\maketitle

\begin{abstract}
Random forests are a statistical learning method widely used in many areas of scientific research because of its ability to learn complex relationships between input and output variables and also its capacity to handle high-dimensional data. However, current random forest approaches are not flexible enough to handle heterogeneous data such as curves, images and shapes. In this paper, we introduce Fréchet trees and Fréchet random forests, which allow to handle data for which input and output variables take values in general metric spaces. To this end, a new way of splitting the nodes of trees is introduced and the prediction procedures of trees and forests are generalized. Then, random forests out-of-bag error and variable importance score are naturally adapted. A consistency theorem for Fréchet regressogram predictor using data-driven partitions is given and applied to Fréchet purely uniformly random trees. The method is studied through several simulation scenarios on heterogeneous data combining longitudinal, image and scalar data. Finally, one real dataset about air quality is used to illustrate the use of the proposed method in practice.
\end{abstract}


\section{Introduction}

Random Forests \citep{breiman2001random} are one of the state-of-the-art machine learning methods.
It owes its success to very good predictive performance coupled
with very few parameters to tune.
Moreover, as a tree-based method, it is able to handle regression and classification (2-class or multi-class)
problems in a consistent manner and deals with quantitative or qualitative
input variables.
Finally, its non-parametric nature allows to proceed high-dimensional data
where the number of input variables is very large in regards of statistical units.

The general principle of a tree predictor is to recursively partition the input space in a binary manner. Starting from the root node which contains all learning observations,
it repeatedly splits each node into two or more child nodes until a stopping
rule is reached. When the input variables are quantitative, splits consist in an input
variable $X^j$ and a threshold $s$, leading to two child nodes containing
observations that verify $\{ X^j \leq s \}$ and $\{ X^j > s \}$ respectively
\citep{cart93}.
For a categorical input variable, a split is a partition of the variable categories into two groups.
The splitting variable as well as the threshold or the categories partition are usually sought to minimize an heterogeneity criterion on child nodes.
The main idea is to partition the input space into more and more homogeneous regions in terms of the output variable.

A limitation of the splitting strategy described above is that all input variables must either live in $\mathbb{R}$ or be categorical, which is not the case with non-Euclidean data such as images, shapes or curves.
As an illustrative example, the real dataset to be analyzed in this paper is from an air quality study made of input variables which are discretely sampled curves representing repeated measurements over time. This example typically corresponds to observations contained in longitudinal (or functional) data. In such settings, the main objective is to predict, for a given observation, its output using the knowledge of inputs trajectories.
If we consider the input data at the trajectory level, then standard ways of splitting nodes cannot be used anymore.
However, ignoring the fact that measurements are repeated observations over time generally leads to an important loss of information for prediction. Thus, one way of analyzing this kind of data is to generalize the notion of split in metric spaces.

Focusing on the framework of functional (or longitudinal) data, some works have
been done to tackle this kind of problem, mainly in the functional data analysis
literature. For instance, \cite{kadri2010nonlinear} introduced a functional
reproducing kernel Hilbert space approach, which go beyond the functional linear
regression \citep{ramsay2005}. Later on, \cite{oliva2015fast} proposed a
nonparametric fast function-to-function regression estimator that uses basis
representation of input and output functions.
In this paper, we focus on the family of tree-based methods, in which several
adaptations to the context of functional data have been proposed.
On one hand, some authors deal with functional outputs (while the inputs are standard).
\cite{doi:10.1080/10618600.1999.10474847} expressed output curves as linear combinations of a spline functions basis and then use multivariate regression tree.
\cite{NERINI20074984} changed the heterogeneity criterion by using Csiszár's $f$-divergence to adapt regression trees to the case where outputs are probability densities.
\cite{10.1145/1143844.1143888} used the kernel trick to project complex outputs onto an Hilbert space to produce a new notion of heterogeneity.
In the context of heterogeneous treatment effects, \cite{pmlr-v130-du21a} proposed to adapt random forests to the problem of estimating the conditional distribution by using the Wasserstein distance between empirical measures. 
On the other hand, some works have been done to deal with functional inputs and standard outputs.
\cite{https://doi.org/10.1002/cem.2849} proposed an adaptation of random forests by using averages of functions values on their respective domains partitions.
\cite{belli2020measure} proposed to extract summaries of functional inputs, (called functional feature extractors).
Finally, in the context of functional inputs and functional outputs, \cite{Brockhaus2017} proposed a regression method called FDboost to fit an additive regression model where each partial effect function is modeled according to a functions basis, such as B-splines. Each partial effect is estimated by a component-wise gradient boosting algorithm. The FDboost method is able to handle function-to-function, function-to-scalar and scalar-to-function regression, which makes it our main competitor in this paper.

In the more general framework of metric spaces, \cite{haghiri2018comparison} proposed an adaptation of random forests in the special case where neither the representation of the data nor the distances between data points are available.
New innovative regression methods have also emerged for the framework of a metric space valued output variable with Euclidean predictors \citep{petersen2019frechet}.
In the present work, we consider the most general possible framework where inputs and outputs lie in metric spaces.

Hence, we consider the framework of a learning sample
$\mathcal{L}_n = \{ (X_1, Y_1), \ldots, (X_n, Y_n) \}$ made of i.i.d.
observations of a pair of random variable $(X, Y) \in \mathcal{X} \times \mathcal{Y}$, where $\mathcal{X}$ is a product of $p$ metric spaces
$(\mathcal{X}_1, d_1) \times \cdots \times (\mathcal{X}_p, d_p)$,
and where $\mathcal{Y}$ is also a metric space with distance $d_{\mathcal{Y}}$.
The main contributions of our work are two-fold: first we introduce a general notion of split in order to tackle inputs that lie in metric spaces.
We define a split as a couple of elements $c_1$ and $c_2$ of a given metric space which allows to build a Voronoï partition, i.e. to separate input elements that are closer to $c_1$ from those closer to $c_2$.
The second contribution is to replace the split criterion in regression trees using the notion of Fréchet variance \citep{frechet1906} in order to cope with outputs in a metric space.
Moreover, to predict outputs we propose to use the Fréchet mean \citep{frechet1906} (which is the natural extension of the standard mean in metric spaces) of the outputs values corresponding to observations belonging to a given tree leaf.
This justifies the names Fr\'echet trees and Fr\'echet random forests hereafter.
Finally, with this generalization of CART trees, Fr\'echet random forests are derived in a rather standard way: a forest predictor is an aggregation of a collection of randomized trees.
In our framework, the aggregation step therefore consists in taking the Fr\'echet mean of individual tree predictions. Therefore, we propose a new class of random forests based on general metric able to take into account various type of data including spatially or timely correlated measurements.

In this paper, we first present the Fr\'echet tree predictor in Section~\ref{sec:Fréchet-trees} before introducing Fr\'echet random forests in Section~\ref{sec:Fréchet-random-forests}. We introduce an extremely randomized version of the Fréchet random forests method in Section~\ref{ERFRF}.
Section~\ref{sec:theory} is dedicated to the analysis of the consistency of Fréchet regressogram estimators using data-driven partitions with output lying in a metric space. We report numerical experiments using simulated longitudinal data to compare our approach with competitive methods, then we analyze two scenarios of heterogeneous data simulations involving curves, images and scalars in Section~\ref{sec:simulations}. An application of Fr\'echet random forests on daily measured air quality data is presented in Section~\ref{sec:app}. Finally, we discuss in Section~\ref{sec:discussion} potential extensions of this work. All the numerical experiments of this paper are reproducible using our R package \texttt{FrechForest}\footnote {\url{https://github.com/Lcapitaine/FrechForest}}.

\section{Fr\'echet Trees}
\label{sec:Fréchet-trees}

\subsection{Fréchet means and Fréchet variance}

The notions of mean and variance are central to the construction of regression trees \citep{cart93}. We introduce in this section the notions of Fréchet empirical mean and Fréchet empirical variance \citep{frechet1948elements}, which are the natural generalization of mean and variance in metric spaces.The use of the  Fr\'echet mean has now become a standard tool for statistical inference from manifold-valued data. For example, it is the key notion allowing to perform PCA for non-Euclidean data such as functional data on Riemannian manifolds (see e.g. \cite{dai2018modeling}, \cite{1318725}, \cite{10.1007/978-3-642-15567-3_4}) or histograms \citep{cazelles:hal-01581699}, and to analyze ensemble of complex objects with their shapes, such as ECG curves \citep{bigot2013} or phylogenetic trees \citep{10.1093/biomet/asx047}.
The methods proposed in this paper allow to perform nonparametric regression between predictors taking their values in different metric spaces and a metric space valued output.

Let $(z_1,\ldots,z_n)$ a sample from a metric space $\left(\mathcal{Z},d\right)$, the empirical Fréchet function is given by
\begin{equation*}
\begin{array}{ccccc}
\mathcal{F}_n & : & \mathcal{Z} & \longmapsto & \mathbb{R}^+ \\
 & & z & \longmapsto & \frac{1}{n}\overset{n}{\underset{i=1}\sum}d^2(z,z_i) \\
\end{array}
\end{equation*}
the function $\mathcal{F}_n(z)$ measures the average squared distance between $z\in\mathcal{Z}$ and $z_1,\ldots,z_n$. We define the empirical Fréchet means $\overline{z}_n$ of the sample $(z_1,\ldots,z_n)$ as any minimizer of the empirical Fréchet function, i.e.
\begin{equation*}
\overline{z}_n\in\underset{z\in\mathcal{Z}}\argmin\ \mathcal{F}_n(z)
\end{equation*}
Note that in general, the Fréchet mean does not always exist and can be non unique. The empirical Fréchet variance $\mathcal{V}_n$ of the sample $(z_1,\ldots,z_n)$ is then given by 
\begin{equation*}
\mathcal{V}_n=\mathcal{F}_n(\overline{z}_n)=\frac{1}{n}\sum_{i=1}^nd^2(z_i,\overline{z}_n)
\end{equation*}
Note that even if the empirical Fr\'echet mean may not be a unique element of the metric space, the Fr\'echet variance is unique.

We give some examples of commonly encountered metric spaces on which the Fréchet mean exists. \cite{doi:10.1137/100805741} prove the existence of the Fréchet mean in the space of probability measures of finite variance with Wasserstein distance. \cite{petersen2019frechet} show the existence and uniqueness of the Fréchet mean in the set of correlation matrices with fixed dimension. \cite{charlier_2013} gives a necessary and sufficient condition for the existence of the Fréchet mean on the unit circle in $\mathbb{R}^2$, he also discusses about the non-uniqueness of the Fréchet mean in such space. General results about the existence and (non-)uniqueness of Fréchet mean on Riemannian manifolds are given in \cite{https://doi.org/10.1112/plms/s3-61.2.371} and \cite{10.1214/aos/1046294456}. Finally, \cite{legouic:hal-01163262} study the existence of Fréchet mean for random probabilities on geodesic space.

Throughout the paper, Fréchet mean and Fréchet variance will always refer to Fréchet empirical mean and Fréchet empirical variance.  For the sake of simplicity, we assume in the rest of the paper that the Fr\'echet mean is unique.

\subsection{Splitting rule}
\label{split}

One key ingredient in the building of a decision tree is the way its 
nodes are split \citep{cart93}.
Splitting a node $t$ of a tree according to some variable $X^{(j)}$ amounts
to find a way of grouping observations of this node into two subsets constituting the
child nodes.
This grouping is usually performed to maximize the differences between the
two resulting child nodes according to the output variable.
However, if variable $X^{(j)}$ is strongly related to the output variable
$Y$, then it is expected that for two observations with ``close'' $X^{(j)}$
values in $(\mathcal{X}_j, d_j)$, associated outputs will be ``close''
in $(\mathcal{Y}, d_{\mathcal{Y}})$.
From this idea, we introduce a way of splitting nodes in general metric spaces. \
Let $\left(\mathcal{Z},d\right)$ be a metric space, a split is any couple of distinct elements $\left(c_1,c_2\right)$ of $\mathcal{Z}$. We define the partition $\mathcal{P}=\left\{P_1,P_2\right\}$ associated with elements $\left (c_1,c_2\right)$ by $P_1=\left\{z\in\mathcal{Z},d\left(z,c_1\right)\leq d\left(z,c_2\right)\right\}$ and $P_2=\left\{z\in\mathcal{Z},d\left(z,c_2\right)< d\left(z,c_1\right)\right\}$.\\ Let $A$ be a subset of the input space $\mathcal{X}$ and for any
$j = 1,\ldots, p$, let $A_j = \{x^{(j)}, x=\left(x^{(1)},\ldots,x^{(p)}\right)\in A\}$ denotes the set of the $j$-th coordinates of the components of A.
Let $\left( c_{j,l}, c_{j,r} \right)$ be a split on $(A_j, d_j)$, denote $A_{j,r}$ and $A_{j,l}$ the right and left child nodes (\emph{i.e.} the associated partition) obtained from the split $\left(c_{j,l}, c_{j,r}\right)$.\\
The quality of the split $\left(c_{j,l}, c_{j,r}\right)$ is then defined by the following measure of Fr\'echet variance decrease:
\begin{equation}
H_{n, j}(A,c_{j,l},c_{j,r}) = \mathcal{V}_n(A) -\frac{N_n\left(A_{j,r}\right)}{N_n\left(A\right)}\mathcal{V}_n(A_{j,r})-\frac{N_n\left(A_{j,l}\right)}{N_n\left(A\right)}\mathcal{V}_n(A_{j,l})  \end{equation} 
where $N_n (A)$, $N_n(A_{j,l})$ and $N_n(A_{j,r})$ are the number of observations of the learning set $\mathcal{L}_n$ belonging to $A$, $A_{j,l}$, $A_{j,r}$ and $\mathcal{V}_n(A),\ \mathcal{V}_n(A_{j,l})$ and $\mathcal{V}_n(A_{j,r})$ are the empirical Fr\'echet variances of outputs in $A$, $A_{j,r}$ and  $A_{j,l}$ \emph{i.e.} \begin{equation*}
\mathcal{V}_n(A)=\frac{1}{N_n(A)}\sum_{i : X_i \in A}
d_{\mathcal{Y}}^2 (Y_i, \overline{Y}_A)\qquad \mbox{(resp. for $A_{j,l}$ and $A_{j,r}$).}
\end{equation*}  $\overline{Y}_{A}$, $\overline{Y}_{A_{j,l}}$ and 
$\overline{Y}_{A_{j,r}}$ are the Fr\'echet means of outputs associated to observations belonging to nodes $A$, $A_{j,l}$ and $A_{j,r}$ \emph{i.e.}
\begin{equation*}
\overline{Y}_{A}= \argmin_{y \in \mathcal{Y}} \sum_{i : X_i \in A} 
d_{\mathcal{Y}}^2 (y, Y_i)\qquad \mbox{(resp. for $\overline{Y}_{A_{j,l}}$ and $\overline{Y}_{A_{j,r}}$).} \end{equation*}
It is worth noting that the decrease in Fr\'echet variance for each possible
split is compared with the output space metric, which makes it
possible to compare splits made on input variables from different metric spaces. 
At last, the split variable $j_n^*$, chosen for splitting  the node is the one that maximizes $H_{n, j}$, that is
\begin{equation}
j_n^* = \argmax_{j \in \{1, \ldots, p\}} H_{n, j} \enspace .
\end{equation} 

It is easy to show that $H_{n,j_n^*} \geq 0$ for all $n$, thanks to the use of the Fr\'echet mean, which means that each split leads to a decrease of Fr\'echet variance.

To determine the successive splits $(c_{j,l},c_{j,r})$, the user defines, in a preliminary step, a split function \emph{i.e.} a way to find the two representatives $(c_{j,l},c_{j,r})$. More precisely, a split function is an application which associates a couple $\left(c_1,c_2\right)\in\mathcal{Z}$ to any sample $\{h_1,\ldots,h_n\}$ from a general metric space $\left(\mathcal{Z},d\right)$. 
For example, the $2$-means algorithm ($k$-means with $k=2$) can be used to determine the representatives. Note that for each metric space $\left(\mathcal{X}_j,d_j\right)$ we can use a different split function. 

\subsection{Tree building}

Starting from the root node (associated with the whole input space
$\mathcal{X}$), nodes are recursively split in order to give a partition of the input space
$\mathcal{X}$.
A node $t$ of the tree is not split if it is pure, that is if the Fr\'echet
variance of this node is null. As a first step in the building process, the tree is developed until all
nodes are pure, leading to the so-called maximal tree.
Then, the pruning algorithm of CART
\citep{cart93} is applied, with the use of the Fr\'echet variance instead of the standard
empirical variance.
At the end of this step, a sequence of nested sub-trees of the maximal tree
is obtained.
Next, the sub-tree associated to the lowest prediction error (estimated by
cross-validation) is selected as the final tree predictor.
The way a Fr\'echet tree predicts new inputs is detailed in the next section.

\subsection{Prediction}

Let $T_n$ be a Fr\'echet tree, we note $\widetilde{T}_n$ the set of leaves (\emph{i.e.}, terminal nodes) of
$T_n$. For each leaf $t \in \widetilde{T}_n$, the Fr\'echet mean of the outputs of observations belonging to $t$ is associated to $t$.
Then the prediction of the output variable associated with any
$x \in \mathcal{X}$ is given by
$\widehat{y} = T_n (x) = \sum_{t\in\widetilde{T}_n} \overline{Y}_t 
\textbf{1}_{x \in t}$,
where $\textbf{1}_S$ denotes the indicator function of a set $S$ and $\overline{Y}_t$ is the Fr\'echet mean of outputs in $t$
$$\overline{Y}_t=\underset{y\in\mathcal{Y}}\argmin \sum_{i:X_i\in t}d_{\mathcal{Y}}^{2}\left(y,Y_i\right)$$
In order to determine to which leaf belongs an observation $x$, it is dropped down the tree as follows.
Starting from the root node, the associated split variable $X^{(j_{1})}$ is considered, together with its two child nodes $A_{j_1,l}$ and $A_{j_1,r}$,
as well as the corresponding representatives $c_{j_1,l}$ and $c_{j_1,r}$.
To decide in which child node $x$ must fall, its $d_{j_1}$-distance
with $c_{j_1,l}$ and $c_{j_1,r}$ must be computed and $x$ goes to
$A_{j_1,l}$ if $d_{j_1} (x^{(j_1)}, c_{j_1,l}) < d_{j_1} (x^{(j_1)}, 
c_{j_1,r})$ and to $A_{j_1,r}$ otherwise.
This process is then repeated until $x$ falls into a leaf.
The error made by $T_n$ on $x$ is defined as:
$$\mathrm{err} (T_n(x)) = d^2_{\mathcal{Y}} (T_n(x), y) \enspace .$$

\section{Fr\'echet random forests}
\label{sec:Fréchet-random-forests}

%
\subsection{An aggregation of Fréchet trees}
A Fr\'echet random forest is derived as standard random forests 
\citep{breiman2001random}: it consists in an aggregation of a collection
of \emph{randomized} Fr\'echet trees.
Here, the same random perturbations as standard random forests
\cite{breiman2001random} are used.
Let $l \in \{1, \ldots, q \}$, consider the $l$-th tree built on a bootstrap sample of the learning sample
$\mathcal{L}_n^{\Theta_l}$ ($n$ observations drawn with replacement among
$\mathcal{L}_n$), the search for the optimized split for each
node of this tree is restricted to a subset of $mtry$ variables randomly drawn among
the $p$ input variables (those random subsets are denoted by $\Theta_l'$ 
hereafter). The $l$-th \textit{randomized} Fr\'echet tree is denoted by $T_n\left(.,\Theta_l,\Theta_l'\right)$ and
can be viewed as a doubly-randomized Fr\'echet tree.
Once all randomized trees are built, the Fr\'echet mean is again used to
aggregated them.
Thus, for any $x\in\mathcal{X}$ the prediction made by the Fr\'echet random
forest is:
$$ \widehat{y} = \argmin_{z \in \mathcal{Y}} \sum_{l=1}^{q} d_{\mathcal{Y}}^2 (z, T_n (x, \Theta_l, \Theta_l') ) \enspace .$$
%
%

\subsection{OOB error and variable importance scores}
\label{sec:oob}

Fr\'echet random forests inherit from standard random forest quantities:
OOB (\textbf{O}ut-\textbf{O}f-\textbf{B}ag) error and variable importance
scores.
The OOB error provides a direct estimation of the prediction error of the
method and proceeds as follows.
The predicted output value, $\widehat{Y}_i^{OOB}$, of the $i$-th observation
$(X_i,Y_i) \in \mathcal{L}_n$, is obtained by aggregating only trees
built on bootstrap samples that do not contain $(X_i,Y_i)$.
The OOB error is then computed as the average squared distance between those 
predictions and the $Y_i$:
$$ \mathrm{errOOB} = \frac{1}{n} \sum_{i=1}^{n} d_{\mathcal{Y}}^2 (Y_i, \widehat{Y}_i^{OOB}) \enspace .$$

Variable importance (VI) provides information on the use of input
variables in the learning task that can be used \emph{e.g.} to perform
variable selection. There are several ways of computing variable importance scores.
Some of them are based on the capacity of given variable to decrease nodes heterogeneity, such as the one already proposed in CART \citep{cart93} or one usually called MDI (Mean Decrease Impurity) in random forests \citep{breiman2001random, louppe2013understanding}.
Another one, sometimes called MDA (Mean Decrease Accuracy) is based on measuring the effect of permuting the values of a given variable, on prediction performance (or accuracy).
In this paper, we generalize the permutation-based VI, because in practice it appears that it suffers less from some selection bias \citep{strobl2007bias, Strobl08, Gre13}.
For $j \in \{1, \ldots, p\}$, variable importance of input variable
$X^{(j)}$, denoted $\mathrm{VI} (X^{(j)})$, is computed as follows.
For the $l$-th bootstrap sample $\mathcal{L}_{n}^{\Theta_l}$, let us define
the associated $\mathrm{OOB}_l$ sample of all observations that were not
picked in $\mathcal{L}_n^{\Theta_l}$.
First, $\mathrm{errOOB}_l$, the error made by tree
$T_n\left(.,\Theta_l,\Theta_l'\right)$ on $\mathrm{OOB}_l$ is computed.
Then, the values of $X^{(j)}$ in the $\mathrm{OOB}_l$ sample are randomly
permuted, to get a disturbed sample $\widetilde{\mathrm{OOB}}_l^j$, and the 
error, $\mathrm{err}\widetilde{\mathrm{OOB}}_{l}^{j}$, made by
$T_n\left(.,\Theta_l,\Theta_l'\right)$ on $\widetilde{\mathrm{OOB}}_l^j$ is
calculated.
Finally, VI of $X^{(j)}$ is defined as:
$$\mathrm{VI}(X^{(j)}) = \frac{1}{q} \sum_{l=1}^{q} \left(
\mathrm{err}\widetilde{\mathrm{OOB}}_{l}^{j} - \mathrm{errOOB}_{l}\right) 
\enspace .$$
%

\subsection{Extremely randomized Fréchet random forests}
\label{ERFRF}
The construction of a Fréchet tree is conditioned by: i) the existence of the Fréchet mean for the output space $\left(\mathcal{Y},d_{\mathcal{Y}}\right)$; ii) the use of a calculable split function for each input space. As mentioned in Section~\ref{split}, in practice the 2-means algorithm can be used as the split function. However, it may not be applicable on all input spaces, for example on input spaces where the Fréchet mean does not exist. In order to have a split function applicable on all input metric spaces, we use the split function introduced by \cite{geurts2006extremely} for regression and classification trees in $\mathbb{R}^p$: let $\texttt{ntry}$ be an integer between 1 and $n(n-1)/2$, we randomly draw $\texttt{ntry}$ different splits \emph{i.e.} $\texttt{ntry}$ different couples of representatives, then we calculate the reduction of the Fréchet variance associated to each of these splits for the response variable and finally we select the split which maximizes the reduction of the Fréchet variance on the response variable. An extremely randomized Fréchet tree (ERFT) is any tree built with this random split function. An aggregation of extremely randomized Fréchet trees is called an extremely randomized Fréchet random forest (ERFRF). Note that when $\texttt{ntry}=n(n-1)/2$ (where $n$ is the sample size of $\mathcal{L}_n$) the node split is no longer random. This splitting strategy has two advantages: it is applicable for any type of input and by taking a low value of $\texttt{ntry}$, it allows to drastically reduce calculation times while having excellent prediction capabilities (see Section~\ref{res1}).

\section{Theory}
\label{sec:theory}
In this section we study the consistency of Fréchet regressogram using data-driven partitions. First, we recall the notions of specific risk and global risk in a general framework before recalling the notion of Fréchet function. Then we remind the notion of family of partitions on $\mathbb{R}^p$. Finally we give the definition of Fréchet regressogram using data-driven partition and a result of its consistency in the case where the input space is $\mathbb{R}^p$ and the output space is a metric space. 

\subsection{Problem}
In this section we present some notations in the general framework where $\mathcal{X}$ is any separable space and $(\mathcal{Y},d)$ is a separable metric space. 
Consider the pair of random variables $\left(X,Y\right)\in\mathcal{X}\times\left(\mathcal{Y},d\right)$. The task is to learn a mapping $\phi:\mathcal{X}\longrightarrow\mathcal{Y}$.\\ For any mapping $\phi:\mathcal{X}\longrightarrow\mathcal{Y}$ the loss function $L$ is given by $$L\left(y,\phi(x)\right)=d^2\left(y,\phi(x)\right)\qquad y\in\mathcal{Y},\ x\in\mathcal{X}$$
The global risk associated with the mapping $\phi$ is defined by \begin{equation}
R(\phi)=\mathbb{E}\left[L(Y,\phi(X))\right]=\mathbb{E}\left[d^2(Y,\phi(X))\right]\end{equation}
The Bayes optimal mapping $\phi^{*}$ is any minimizer of the global risk function \emph{i.e.}\begin{equation}
\phi^{*}\in\underset{\phi:\mathcal{X}\longrightarrow\mathcal{Y}}\argmin\ R(\phi)
\end{equation}
When $\mathcal{X}$ and $\mathcal{Y}$ are separable, according to \cite{blackwell1984factorization} the global risk can be factorized as
\begin{equation}
R(\phi)=\mathbb{E}_{X}\left(\mathbb{E}_{Y}\left[d^2(Y,\phi(X))|X\right]\right)
\end{equation}
We define the point risk function of $\phi$ by \begin{equation}
r\left(x,\phi(x)\right)=\mathbb{E}_{Y}\left[L(Y,\phi(X))|X=x\right]=\mathbb{E}_{Y}\left[d^2(Y,\phi(X))|X=x\right]
\end{equation}
The Bayes optimal point-risk mapping $\phi^{*}$ is defined by
\begin{equation}\phi^{*}(x)\in\underset{y\in\mathcal{Y}}\argmin\ r\left(x,y\right), \quad \mbox{where} \quad  r\left(x,y\right)= \mathbb{E}_Y\left[d^2(Y,y) |X=x \right].
\end{equation}
This mapping introduced in \cite{petersen2019frechet} is called Fréchet regression function. 

\subsection{Family of partitions}
Let $\mathcal{X}=\mathbb{R}^p$, denote $\mathcal{Z}=\mathbb{R}^p\times\mathcal{Y}$ and let $\pi_n$ be a partitioning rule of $\mathbb{R}^p$ \emph{i.e} a function that associates a measurable partition of $\mathbb{R}^p$ to any vector $\left(z_1, \ldots, z_n\right)\in\mathcal{Z}^n$. We note $\mathcal{A}_n$ the family of all the partitions we can obtain with $\pi_n$:
\begin{equation}
\mathcal{A}_n:=\{\pi_n\left(z_1,\ldots,z_n\right), \left(z_1,\ldots,z_n\right)\in\mathcal{Z}^n\}
\end{equation}
We denote $\mathcal{C}\left(\mathcal{A}_n\right)=\underset{\pi\in\mathcal{A}_n}{\sup}|\pi|$ the maximal number of cells for the partitions family $\mathcal{A}_n$. Finally, let $\mathcal{A}$ be a family of partitions, let $x_1,\ldots,x_n$ $n$ points of $\mathbb{R}^p$ and let $B=\left\{x_1,\ldots,x_n\right\}$. We note $\Delta\left(\mathcal{A},x_1^n\right)$ the number of distinct partitions $$\left\{A_1\cap B,A_2\cap B, \ldots, A_n\cap B\right\}$$ induced by the partitions $\{A_1,\ldots,A_n\}\in\mathcal{A}$. The growing function of the partitions family $\mathcal{A}$ is defined by
\begin{equation}
\Delta_n^*\left(\mathcal{A}\right)=\underset{x_1^n\in\mathbb{R}^{d.n}}{\max}\Delta\left(\mathcal{A},x_1^n\right)
\end{equation}
Let $\left(X_1,\ldots,X_n\right)$ a sample made of independent observations with the same distribution as $X$. Denote $\mu$ the distribution of $X$ and $\mu_n$ the empirical distribution of the sample $\left(X_1,\ldots,X_n\right)$. The following Lemma can be found in \cite[lemma 1]{lugosi1996consistency}. 

\begin{lemma}
\label{lem2}
Let $\mathcal{A}$ be any collection of partitions of $\mathbb{R}^p$. For every $n\geq 1$ and every $\epsilon>0$,

\begin{equation}
\mathbb{P}\left(\underset{\pi\in\mathcal{A}}\sup\underset{A\in\pi}\sum \left|\mu(A)-\mu_n(A)\right|>\epsilon\right)\leq 4\Delta_n^*\left(\mathcal{A}\right)2^{\mathcal{C}\left(\mathcal{A}\right)}\exp\left(-n\epsilon^2/32\right)
\end{equation}
\end{lemma}

\subsection{Fréchet regressogram}
Let $\mathcal{L}_n=\left\{(X_1,Y_1),\ldots,(X_n,Y_n)\right\}$ be a learning sample made of independent observations with same distribution as $\left(X,Y\right).$ Let $\pi_n$ a partitioning rule, we define the Fréchet regressogram estimator by\begin{equation}
\label{eq:est}
T_n\left(x\right)=\underset{y\in\mathcal{Y}}\argmin\ \frac{1}{n}\sum_{i=1}^{n}d^2(Y_i,y)\textbf{1}\{X_i\in \pi_n[x]\}
\end{equation}
where $\pi_n[x]$ denotes the unique cell containing $x$. The goal is then to show that under certain assumptions on the metric space $(\mathcal{Y},d)$, on the distribution of $(X,Y)$ and on the partitioning rule, this estimator is consistent for the point risk as well as for the global risk. 

We recall the definitions of doubling dimension and covering numbers given in \cite{GOTTLIEB2016105}.

\begin{definition}[Doubling dimension]
Let $(\mathcal{Y},d)$ be a metric space, let $\lambda_{\mathcal{Y}}>0$ be the smallest positive integer such that every ball in $\mathcal{Y}$ can be covered by $\lambda_{\mathcal{Y}}$ balls of half its radius. The doubling dimension of $(\mathcal{Y},d)$ is then defined as $\ddim(\mathcal{Y}):=\log_2(\lambda_{\mathcal{Y}})$.
\end{definition}

\begin{definition}[Covering numbers]
The $\epsilon$-covering number $\mathcal{N}(\epsilon,\mathcal{Y},d)$ of a metric space $(\mathcal{Y},d)$ is defined as the smallest number of balls of radius $\epsilon$ that suffices to cover $\mathcal{Y}$.
\end{definition}
The diameter of a metric space $(\mathcal{Y},d)$, denoted $\diam(\mathcal{Y})$ is defined by $\diam(\mathcal{Y})=\underset{y_1,y_1\in\mathcal{Y}}\sup d(y_1,y_2)$. When both the diameter and doubling dimension of the metric space $(\mathcal{Y},d)$ are finite, according to \cite{GOTTLIEB2016105}, the following lemma allows to bound the $\epsilon-$covering number. 
\begin{lemma}
\label{lem1}
Let $\left(\mathcal{Y},d\right)$ be a metric space with finite diameter $\diam(\mathcal{Y})<\infty$ and finite doubling dimension $\ddim(\mathcal{Y})<\infty$. Then, for every $0<\epsilon\leq \diam(\mathcal{Y})$
\begin{equation}
N(\epsilon,\mathcal{Y},d)\leq \left(\frac{2\diam(\mathcal{Y})}{\epsilon}\right)^{\ddim(\mathcal{Y})}
\end{equation}

\label{L1}
\end{lemma}
We now state the main result of our analysis.
\begin{theorem}
Let $\left(\mathcal{Y},d\right)$ with finite diameter $\diam(\mathcal{Y})$ and finite doubling dimension $\ddim\left(\mathcal{Y}\right)$.  Let $\pi_n$ be a partitioning rule on $\mathbb{R}^p$, $\Pi_n$ be the family of partitions of $\mathbb{R}^p$ obtained from $\pi_n$ and $\mathcal{V}_n[x]=\mathbb{E}(\Vol(\pi_n[x]))$ be the expected volume of the cell containing $x$. Assume that the following properties hold: 
\begin{enumerate}
\item[\textbf{P1.}]\label{P1} We assume that $\left(X,Y\right)$ has uniformly continuous and bounded density $\rho$ and the marginal $\rho_{X}$ verifies $0<\rho_{\min}\leq\rho_X$
\item[\textbf{P2.}] $\frac{\mathcal{C}(\Pi_n)}{n}\rightarrow 0$
\item[\textbf{P3.}]  $\frac{\log(\Delta_n^*(\Pi_n))}{n}\rightarrow 0$
\item[\textbf{P4.}] $\frac{\log \mathcal{V}_n[x]}{n}\rightarrow 0$
\item[\textbf{P5.}] $\frac{1}{\mathcal{V}_n[x]}=o(\frac{n}{\log n})$
\item[\textbf{P6.}] $\diam(\pi_n[x])\rightarrow 0$ almost surely
\end{enumerate}
 then 
\begin{equation}
\label{eq:conv}
\lim\limits_{n\rightarrow\infty}\left|r\left(x, T_n(x)\right)-\underset{y\in\mathcal{Y}}\min\ r\left(x,y\right)\right|=0, \qquad\mbox{a.s.}
\end{equation}
Furthermore, 
\begin{equation}
\label{eq:convmoy}
\lim\limits_{n\rightarrow\infty} R(T_n)-R(\phi^*)=0,\qquad \mbox{a.s}
\end{equation}
\label{T1}
\end{theorem} 
\begin{proof}
The proof can be found in Appendix~\ref{AN1}.
\end{proof}

\subsection{Fréchet purely uniformly random trees}
In this (sub)section the input space considered is $\mathcal{X}=[0,1]$. As several theoretical works on regression trees, we consider a simplified version of Fréchet trees. Hence, we study a variant of the purely random trees introduced in \cite{doi:10.1080/10485252.2012.677843}, denoted Fréchet purely random tree.  

\begin{definition}[Fréchet purely uniformly random tree]
Let $\mathcal{L}_n=\{(X_1,Y_1),\ldots,(X_n,Y_n)\}$ be a learning sample of i.i.d measurements in $[0,1]\times(\mathcal{Y},d)$. Let $k_n$ be a positive integer and $U_1,\ldots,U_{k_n}$ be $k_n$ i.i.d uniformly drawn random variables on $[0,1]$. Denote $U_{(1)},\ldots,U_{(k_n)}$ the order statistics, the Fréchet purely random tree predictor $FPURT_n$ is given by 
\begin{equation}
FPURT_n(x)=\underset{y\in\mathcal{Y}}\argmin\frac{1}{n}\sum_{j=0}^{k_n}\sum_{i=1}^nd^2(y,Y_i)\textbf{1}\{U_{(j)}\leq x\leq U_{(j+1)}\}\qquad \forall x\in[0,1]
\end{equation}
with $U_{(0)}=0$ and $U_{(k_n+1)}=1$
\end{definition}

\begin{corollary}
Let $\left(\mathcal{Y},d\right)$ with finite diameter $\diam(\mathcal{Y})$ and finite doubling dimension $\ddim\left(\mathcal{Y}\right)$. Let $k_n$ be an integer depending on $n$. Assume the following assumptions:
\begin{enumerate}
\item[\textbf{A1.}]\label{A1} We assume that $\left(X,Y\right)$ has uniformly continuous and bounded density $\rho$ and the marginal $\rho_{X}$ verifies $0<\rho_{\min}\leq\rho_X$
\item[\textbf{A2.}]\label{A2} $k_n\rightarrow \infty$ as $n\rightarrow \infty$ and $k_n=o(n/\log n)$
\end{enumerate}
 hold then the Fréchet purely uniformly random tree estimator is consistent for the global risk i.e \begin{equation}
\lim\limits_{n\rightarrow\infty} R(FPURT_n)-R(\phi^*)=0,\qquad \mbox{a.s}
\end{equation}
\end{corollary}

\begin{proof}
Let $\pi_n$ be the partitioning rule used to build $FPURT_n$ and let $\Pi_n$ the family of partitions associated with $\pi_n$. The interval $[0,1]$ is partitioned into $k_n+1$ intervals, then $\mathcal{C}(\Pi_n)=k_n+1$ which implies $$\frac{\mathcal{C}(\Pi_n)}{n}=\frac{k_n+1}{n}\underset{n\longrightarrow \infty}\rightarrow 0$$
It is easy to show that $\Delta_n^*(\Pi_n)\leq n^{k_n}$, then we deduce from $k_n=o(n/\log n)$ that $$\frac{\log \Delta_n^*(\Pi_n)}{n}\leq \frac{k_n\log n}{n}\underset{n\rightarrow \infty}\longrightarrow 0$$
From \cite{arlot2014analysis} (page 34-36) we have that the expected volume (diameter in dimension one) of the interval containing $x$ is:\begin{equation}
\mathcal{V}_{n}[x]=\frac{2-x^{k_n+1}-(1-x)^{k_n+1}}{k_n+1}\qquad \forall x\in [0,1]
\end{equation}
Hence, $\mathcal{V}_n[x]\leq \frac{2}{k_n+1}$, then \begin{equation*}
\frac{\log \mathcal{V}_{n}[x]}{n}\leq \frac{\log 2 - \log(k_n+1)}{n}\underset{n\rightarrow\infty}\longrightarrow 0
\end{equation*}
Finally, for $x\in\{0,1\}$ \begin{equation*}
\frac{\log n}{n\mathcal{V}_{n}[x]}=\frac{(k_n+1)\log n}{n}\underset{n\rightarrow\infty}\longrightarrow 0
\end{equation*}
and for every $0<x<1$
\begin{equation*}
\frac{\log n}{n\mathcal{V}_{n}[x]}=\frac{(k_n+1)\log n}{(2-x^{k_n+1}-(1-x)^{k_n+1})n}\underset{n\rightarrow\infty}\sim \frac{(k_n+1)\log n}{2n}\underset{n\rightarrow\infty}\longrightarrow 0
\end{equation*}
We demonstrated that the properties \textbf{P1}-\textbf{P5} of Theorem~\ref{T1} are verified. We thus conclude that the one dimension $FPURT_n$ estimator is point-wise consistent as well as consistent for the global risk.
\end{proof}

Here we considered purely uniformly random trees in dimension 1. \cite{arlot2014analysis} defined purely random trees in $\mathbb{R}^p$. Even if it may be possible to apply Theorem~\ref{T1} to these trees, it appears far more difficult to get the probability distribution of the volume of the cell containing $x$, and thus in verifying the properties $\textbf{P4}$ and $\textbf{P5}$. Indeed, as soon as we consider $p>1$, the recursive character of the cuts makes the calculations much more complex. This problem is out of the scope of this paper.

\section{Simulation study}
\label{sec:simulations}

In this section, we study the behavior of Fréchet random forests through two simulation scenarios.

\subsection{First scenario, longitudinal data}

\subsubsection{Two temporal behavior functions scheme}

The first scenario deals with the analysis of longitudinal data where inputs and outputs are curves. We simulate $n=100,\ 200,\ 400$ and $1000$ observations of $p=6$ input variables according to the following model for any $i = 1, \ldots, n$ and for any $j\in\{1,\ldots,6\}$:
\begin{equation}
X_i^{(j)} (t) = \begin{cases} \beta_i \left( f_{j,1} (t) \textbf{1}_{\{G^j_i=1\}} +
f_{j,2} (t) \textbf{1}_{\{G^j_i=2\}}\right) + W_i^1 (t) \quad \mbox{if }j\in\{1,2\}\\
\beta_i' \left( f_{j,1} (t) \textbf{1}_{\{G'^j_i=1\}} +
f_{j,2} (t) \textbf{1}_{\{G'^j_i=2\}}\right) + W_i^1 (t)\quad \mbox{if }j\in\{3,4,5,6\}\end{cases}
\label{simX}
\end{equation}
where $X^{(j)}_i(t)$ is the observation of the $j$th input variable at time $t$ for the $i$th curve (individual); $t$ browses a regular subdivision of $[0,1]$ with a step size of 0.05, $G^j_i$ and $G'^j_i\sim \mathcal{U}\left(\left\{1,2\right\}\right)$, $\beta_i$ and $\beta_i'\sim\mathcal{N}\left(1,0.3\right)$, $W_i^1(t)$ is a
Gaussian white noise with standard deviation $0.02$ and $f_{j,1}$ and $f_{j,2}$ are
defined as follows:
\begin{equation*}
\begin{cases}
			f_{1,1}(t) = 0.5t + 0.1 \sin(6t)  \\
			f_{1,2}(t) = 0.3 - 0.7 (t-0.45)^2 \\
			f_{2,1}(t) = 2 (t - 0.5)^2 - 0.3 t \\
			f_{2,2}(t) = 0.2 - 0.3t + 0.1 \cos(8t) \\
			f_{3,1}(t)=f_{1,1}(t) \\
			f_{3,2}(t)=f_{1,2}(t)
\end{cases}
\begin{cases}
			f_{4,1}(t)=f_{2,1}(t) \\
			f_{4,2}(t)=f_{2,2}(t) \\
			f_{5,1}(t)= 0.5t^2-0.15\sin(5t)\\
			f_{5,2}(t)=0.5t^2\\
			f_{6,1}(t)=0.6\log(t+1)-0.3\sin(5t)\\
			f_{6,2}(t)=0.6\log(t+1)+0.3\sin(5t)
\end{cases}
\label{eq:tempoBehav}
\end{equation*}

\begin{figure}
\centering
\includegraphics[width=\textwidth, height=0.3\textheight]{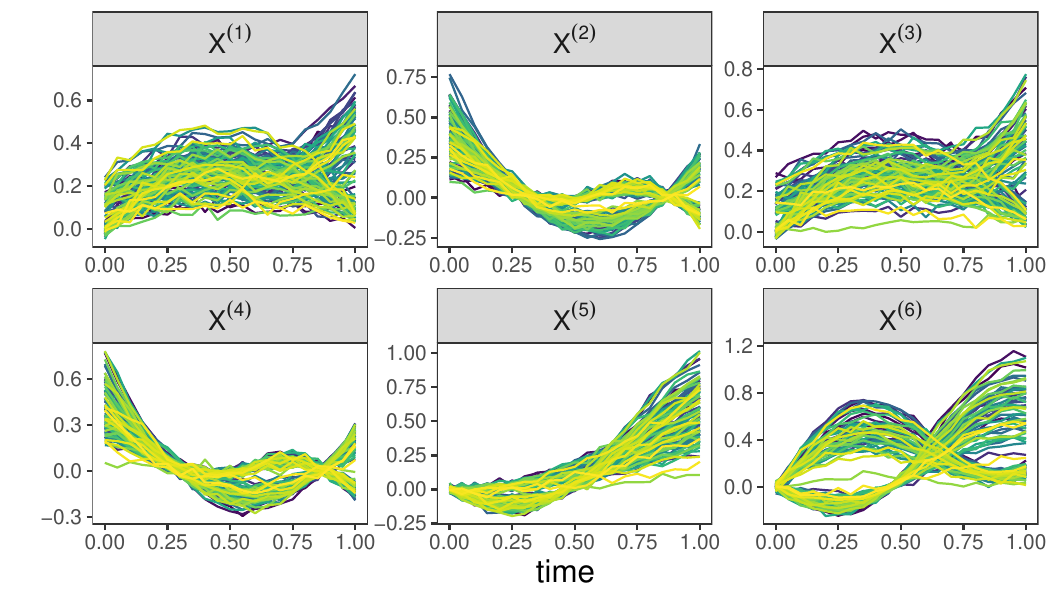}
\caption{Dynamics of $n=100$ simulated input trajectories according to the model~\eqref{simX}}. 
\label{fig:compX}
\end{figure}

The terms $G^j_i$ and $G'^j_i$ allow to randomly affect typical temporal behaviors, defined by 
$f_{j, 1}$ and $f_{j, 2}$ functions, to observations.
The $\beta_i$ and $\beta'_i$ are dilatation/shrinkage terms of $f_{j, 1}$ or $f_{j, 2}$, while $W_i^1(t)$ corresponds to an additive noise. As illustrated in Figure~\ref{fig:compX}, for each input variable, the observed trajectories are variations of the typical temporal behavior functions. The observations are divided into two groups of trajectories. 

Output variable $Y$ is simulated in a similar way. The pair $(G_i^1,G_i^2)$ is used
to determine a trajectory for the output variable, this is the primary link between $X_i$ and $Y_i$
\begin{equation}
Y_i(t) = \beta_i \sum_{j=1}^2 \sum_{k=1}^2 g_{j,k}(t) \textbf{1}_{\{G^j_i=j\}} 
\textbf{1}_{\{G_i^k=k\}} + W_i^2(t)
\label{simY}
\end{equation}
where $Y_i(t)$ is the $i$th output curve measured at time $t$; $t$ browses the same subdivision as in \eqref{simX}, $\beta_i$ are the same coefficients used in \eqref{simX}, $W_i^2\left(t\right)$ is a Gaussian white noise with standard deviation $0.05$ and $g_{j,k}$ are given by: 
\begin{equation}
\left\{
\begin{array}{l@{\hskip 5ex}l}
            g_{1,1}(t) = t + 0.3 \sin(10\left(t+1\right))\\
            g_{1,2}(t) = t+2(t-0.7)^2 \\
            g_{2,1}(t) = 1.5\exp\left(-\frac{(t-0.5)^2}{0.5}\right)- 0.1\left(t+1\right)\cos(10t)\\
            g_{2,2}(t) = \frac{\log(13(t+0.2))}{1+t} \\
\end{array}
\right.
\label{eq:behavY}
\end{equation}

The response curves are distributed according to four different trajectory shapes, one for each pair of possible trajectory shapes for the first two input curve variables $X^{(1)}$ and $X^{(2)}$. Of note, the variables $X^{(3)}$ and $X^{(4)}$ are simulated using the same temporal functions as variables $X^{(1)}$ and $X^{(2)}$; however, the trajectories of variables $X^{(3)}$ and $X^{(4)}$ are simulated from $G'^j$ and not from $G^j$ and thus have no relation with the output variable $Y$.

\subsubsection{Three temporal behavior functions scheme}

\begin{figure}
\centering

\begin{tikzpicture}
\node (tb) at (0,3) {2 temporal behaviors functions};
\node (CSX1) at (-4,0) {\includegraphics[width=0.25\textwidth, height=0.18\textheight]{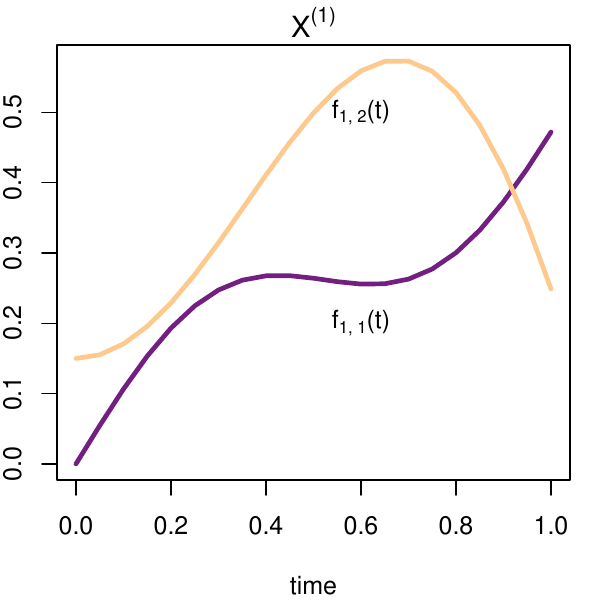}};
\node (CSX2) at (0,0) {\includegraphics[width=0.25\textwidth, height=0.18\textheight]{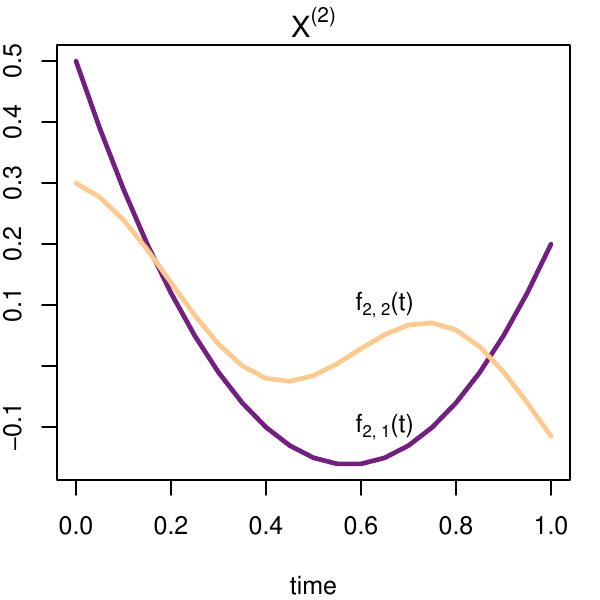}};
\node (YC22) at (4,0) {\includegraphics[width=0.25\textwidth, height=0.18\textheight]{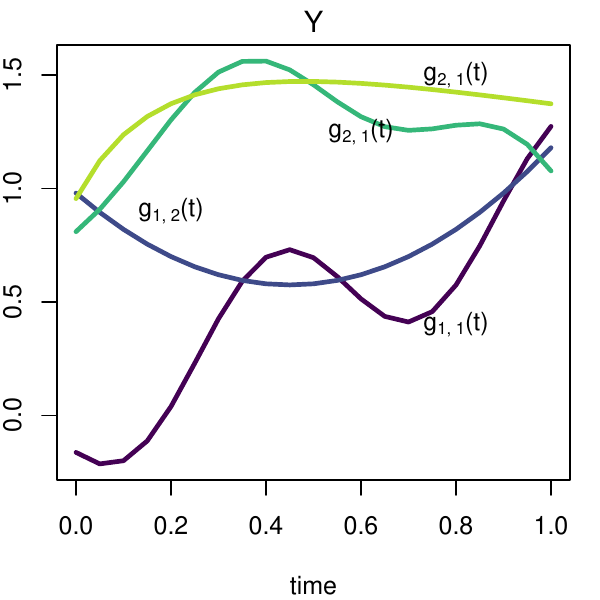}};

\node (tb) at (0,-3) {3 temporal behaviors functions};
\node (CCX1) at (-4,-6) {\includegraphics[width=0.25\textwidth, height=0.18\textheight]{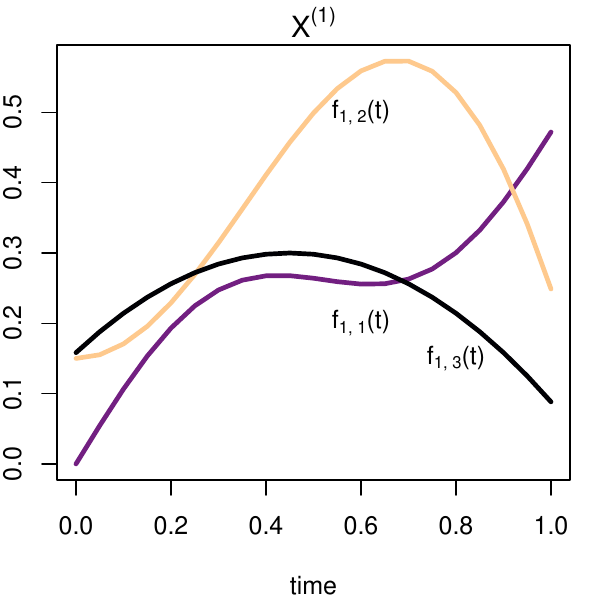}};
\node (CCX2) at (0,-6) {\includegraphics[width=0.25\textwidth, height=0.18\textheight]{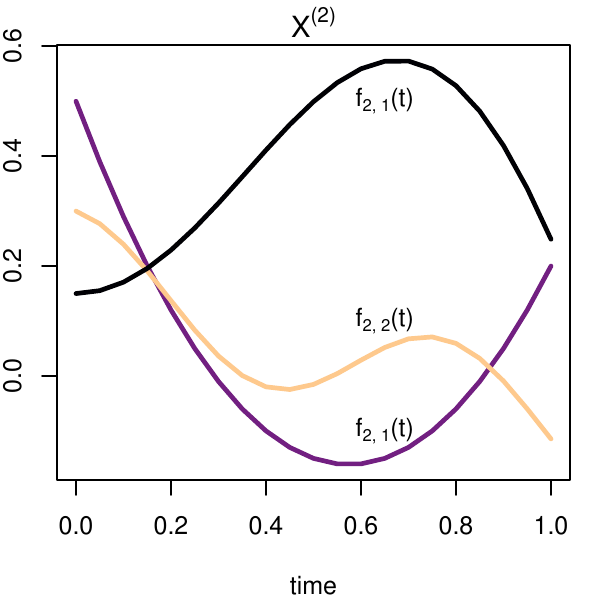}};
\node (YC33) at (4.5,-6) {\includegraphics[width=0.33\textwidth, height=0.188\textheight]{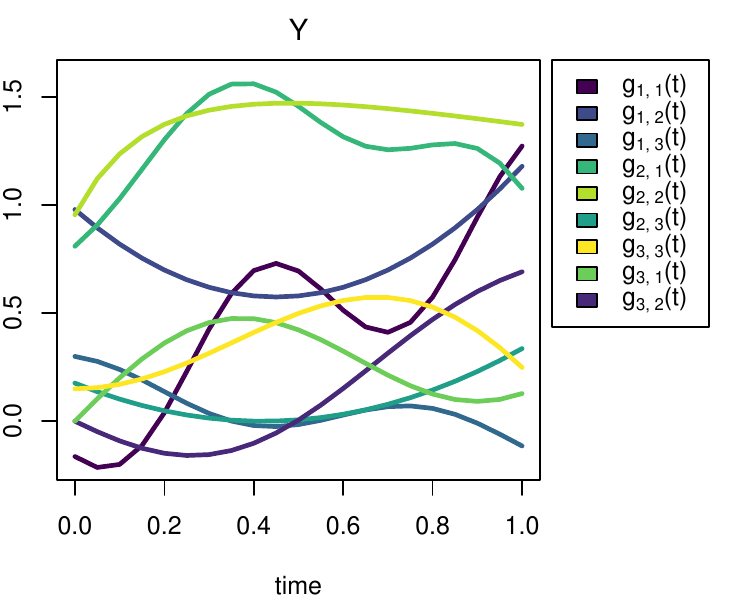}};

\node (tb) at (0,-9) {50 simulated curves according to the 3 temporal behaviors functions scheme};
\node (CfullX1) at (-4,-12) {\includegraphics[width=0.25\textwidth, height=0.18\textheight]{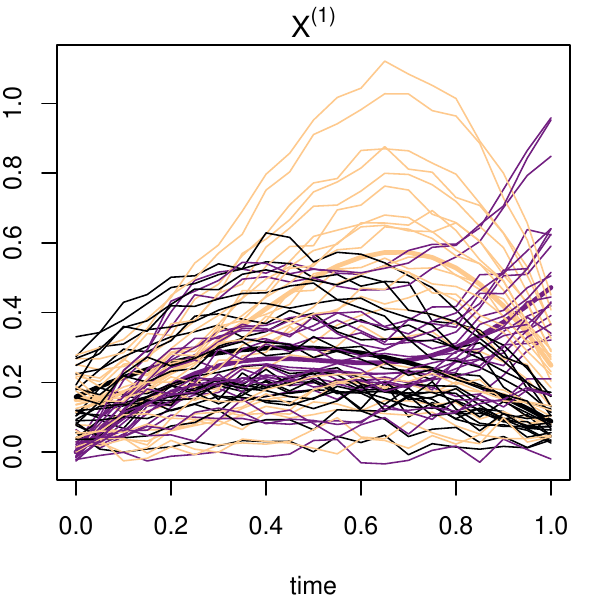}};
\node (CfullX2) at (0,-12) {\includegraphics[width=0.25\textwidth, height=0.18\textheight]{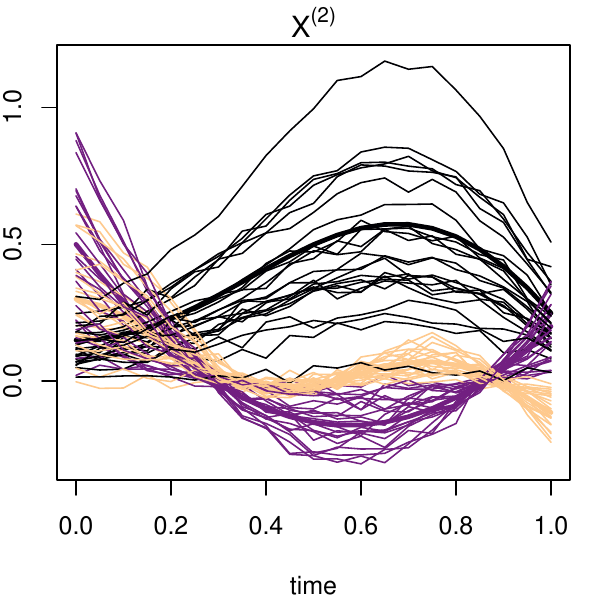}};
\node (CfullY) at (4,-12) {\includegraphics[width=0.25\textwidth, height=0.18\textheight]{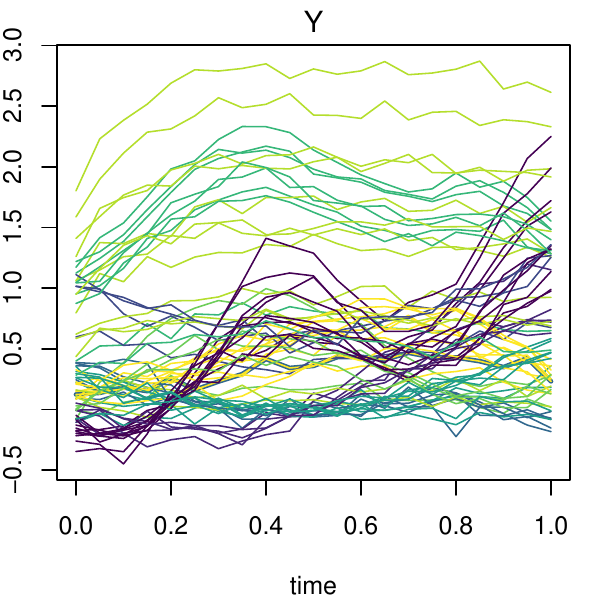}};

\end{tikzpicture}

\caption{The first two lines show the time behavior functions for schemes \eqref{simX} and \eqref{simX3}, for the first two input variables $X^{(1)}$ and $X^{(2)}$ and the output $Y$. The third row shows 50 simulated dynamics according to scheme \eqref{simX3} (see Appendix~\ref{Complement3}).}
\label{fig:confounded}

\end{figure}

This simulation scheme is a variation of the previous one. Here, for each explanatory variable, the curves are simulated from 3 time behavior functions instead of 2 which leads the output curves to be constructed according to 9 different mean behavior functions that are either expanded or contracted. Details about the additional temporal behavior functions can be found in Appendix~\ref{Complement3}. \\
In this simulation scheme, we consider 3 groups of curves instead of 2 to make the task of splitting by Fréchet trees more difficult, see Figure~\ref{fig:confounded} for an illustration of the differences between the schemes with 2 and 3 time behavior functions for the first two input variables and the output.

The objective of these first two scenarios is first to compare the predictive capabilities of the introduced methods with the standard ones such as Breiman's CART trees and random forests, linear mixed-effects models but also to compare with the functional method FDboost \citep{Brockhaus2017}. In a second step, we will focus on the flexibility of the Fréchet random forests, in particular on the stability of the prediction error as well as of the importance scores of the variables in two situations frequently encountered in practice: first when we have time shifts on curves and second when we have missing data.

\subsection{Second scenario, predict curves with images, scalars and curves}

In this scenario, we want to predict output curves from inputs that are curves, scalars and images to illustrate the flexibility of the Fréchet RF method, in particular its ability to learn about different types of inputs and outputs. The input curve variables are simulated according to the model~\eqref{simX} of the first scenario with $\beta_i$ and $\beta_i'$ drawn according to $\mathcal{N}(1,1)$ in order to have large variations of the curves around their average temporal behavior. Similarly, the output curves are simulated according to the model~\eqref{simY} of the first scenario. Let $\left(\mathcal{M}^1_i\right)_i$ and $\left(\mathcal{M}^2_i\right)_i$ two sequences of handwritten images of numbers 1 (for $\mathcal{M}^1_i$) and 2 (for $\mathcal{M}^2_i$) randomly drawn from the MNIST dataset \citep{lecun-mnisthandwrittendigit-2010}. We simulate two input image variables $I^{(1)}$ and $I^{(2)}$ according to the following model: 
\begin{equation}
I^{(j)}_i=\mathcal{M}^1_i\textbf{1}_{\{G^j_i=1\}} + \mathcal{M}^2_i\textbf{1}_{\{G^j_i=2\}}\quad \mbox{for}\ j\in\{1,2\};\ i\in\{1,\ldots,n\}
\end{equation}
where $G^j_i$ are the same draws as those used to simulate the input and output curves in model~\eqref{simX} and model~\eqref{simY}. Finally, consider the two real input variables $R^{(1)}_i=\beta_i$ and $R^{(2)}_i=\beta_i'$, where the $\beta_i$ and $\beta_i'$ are the same as those used to simulate the input and output curves. The first variable $R^{(1)}$ determines the intensity of the contraction/expansion of the $X^{(1)}$ and $X^{(2)}$ response curves. It is important to note that the link between the output curves and the input variables is entirely contained in the pairs $(G^1_i,G_i^2)$ which determine the general shape of the output curve as well as the $\beta_i$ which determine the compression/expansion of the output curves. The pairs $(G^1_i,G_i^2)$ as well as the $\beta_i$ are used to simulate the first two curve input variables $X^{(1)}$ and $X^{(2)}$. However the two image variables are constructed only from the pairs $(G^1_i,G_i^2)$ and the scalar variables are $\beta_i$ and $\beta_i'$. \\
In this scenario, we are mainly interested in the ability of random forests to handle heterogeneous data and to extract information from variables of different natures. \\
A third simulation scenario in which images from the MNIST dataset \citep{lecun-mnisthandwrittendigit-2010} are predicted from curves is presented in the Appendix~\ref{app:AN2}.

\subsection{Results}

\subsubsection{First scenario}
\label{res1}

\paragraph{Distance and split function choices for Fréchet trees and forest}
First, we need to determine a metric for each input and output space. In the case of longitudinal data \emph{i.e.} when repeated measurements of quantitative variables are available over time,  the observations of $p$ input and one output variables can thus be represented by time-dependent curves. In this case, the $i$-th observation $X_i$ is a curve from $\mathcal{I}_1 \times 
\cdots \times \mathcal{I}_p \subset \mathbb{R}^{p}_{+}$ to $\mathbb{R}^{p}$ (where $\mathcal{I}_1=[0,1]$ and $\mathcal{I}_2=[0,1]$ in the first scenario), and
$Y_i$ is a curve from $\mathcal{J} \subset \mathbb{R}_{+}$ to $\mathbb{R}$.
We choose to equip the resulting curves spaces with the Fr\'echet distance $d_\mathcal{F}$ introduced in \cite{frechet1906} defined for two real-valued curves $f$ and $g$ with support in 
$\mathcal{I} \subset \mathbb{R}_{+}$ as
$$ d_{\mathcal{F}} (f, g) = \inf_{\alpha, \beta} \max_{t \in \mathcal{I}}
| f(\alpha(t)) - g(\beta(t)) | $$
where $\alpha$ and $\beta$ are any re-parameterizations of $\mathcal{I}$. The definition is the same in the discrete case (polygonal curves), except that $t$ takes values on $\mathcal{I}$ by intervals, see \cite{doi:10.1142/S0218195995000064} for a full description of Fr\'echet distance for discretely sampled curves. This distance is a natural measure of similarity between the shapes of curves and has been widely used in various applications such as signature authentication
\citep{10.1007/978-3-540-92137-0_51}, path classification \citep{10.1371/journal.pone.0150738} and speech
recognition \citep{kwong1998parallel}. Note that, unlike several classical distances, the calculation of the Fr\'echet distance does not require the same number of measurements, nor the same observations times on the two trajectories. 
Once we have determined the metrics used for the different spaces we need to define the split function used to cut on the input spaces. The 2-means algorithm 
for longitudinal data using Fr\'echet distance and Fr\'echet mean introduced in \cite{10.1371/journal.pone.0150738} is chosen on each input space to determine the different competing splits. 
This split function called \texttt{kmlShape} is an adaptation of the $k$-means method tailored to one-dimensional curves. It allows to find groups of trajectories based on their 
shapes (which are usually not found by conventional methods, \emph{e.g.} based on Euclidean distance).\\

\paragraph{Computational complexity} We now analyze the computational complexity of our method in this framework. We note $n_t$ the number of measurement times per individual (which we consider at first to be the same for each individual). According to \cite{10.1371/journal.pone.0150738}, the algorithmic complexity of the split function we used is of order $\mathrm{O}(2\times n\times n_t^2)$ where $n$ is the number of curves. Still according to \cite{10.1371/journal.pone.0150738}, the computation of the Fréchet distance between two curves of size $n_t$ is of order $\mathrm{O}(n_t^2)$ and the approximation of the Fréchet mean of $n$ curves is of order $\mathrm{O}\left(n\times n_t^2\right)$. At each split, the computation of the initial Fréchet variance (before splitting) as well as that in each child node is necessary. We then deduce that for each splitting, the computation of the decrease of the Fréchet variance is of the order $\mathrm{O}(n\times n_t^2 + n\times n_t^2)$. In the case of a Fréchet tree, the split function is applied to the $p$ input variables. Finally, the overall complexity of a split for a Fréchet tree on curves as inputs and output is of order $\mathrm{O}\left(p\times n\times n_t^2\right)$. Recall that the complexity of a standard CART tree splitting for $n$ observations and $p$ input variables is given by $\mathrm{O}\left(p\times n\times \log n\right)$. Thus, in the longitudinal framework this complexity becomes $\mathrm{O}\left(p\times n\times n_t\times \log (n\times n_t) \right)$. Indeed, as we consider the case where inputs curves and output curves are all observed at the same time points, each line (and thus each observation) is one measurement for one individual and we have therefore $n\times n_t$ independent observations in total for an unchanged number of input variables. We then notice that splitting  a node with a Fréchet tree has a computational complexity lower than the one of a standard CART tree when $n_t\leq \log (n\times n_t)$ and thus $\frac{e^{n_t}}{n_t}\leq n$ \emph{i.e.} when the number of individuals $n$ grows exponentially with respect to the number of time measurements $n_t$.\\ For instance, in a longitudinal framework where each individual would have 10 measurement times, from the moment the number of individuals exceeds 2202, the Fréchet tree method would be faster than the CART tree method. It remains true for the associated random forests which have complexities that derive directly from those of the splitting. It is important to note that in our implementation of the methods presented in this paper in the form of the R package \texttt{FrechForest} the emphasis has been put strongly on the flexibility of the possible inputs and outputs and not on the speed of execution. Another package\footnote {available at \url{https://github.com/Lcapitaine/ExtraFrech.jl}} currently under development in the Julia language focuses entirely on execution time but requires a more rigid structure for this. \\

\paragraph{Competing methods} Fr\'echet trees and Fr\'echet random forests were compared on simulated datasets to standard CART trees \citep{cart93} and standard random forests \citep{breiman2001random} as well as standard existing methods for longitudinal data analysis such as linear mixed effects model (LMEM) with a random intercept and a random effect on time and the boosting functional regression method FDboost \citep{Brockhaus2017} with optimized number of iterations. FDboost was considered because it is a flexible functional boosting method that is able to handle regression problems with functions as well as scalars as inputs and outputs. By its nature and flexibility this method is a natural competitor to the methods introduced in this paper. 
As explained in the previous paragraph on the computational complexity, for the standard CART and RF methods each observation time of an output curve is related to the corresponding time of the input curves. Thus, all measurements of the same individual are considered independent by these methods, which is a fundamental difference from our approach. 

The prediction errors (mean squared error) of all the methods are estimated on several sample sizes $n=100,\ 200,\ 400$ and 1000 using for each sample size, $100$ datasets simulated according to models \eqref{simX} and \eqref{simY}. For each simulated dataset $\mathcal{L}_n$, we randomly divide $\mathcal{L}_n$ into a training set (with $0.8 n$ observations) and a test set (made of the remaining $0.2 n$ observations). The Fréchet distance is used on the curved input and output spaces to build Fréchet trees and Fréchet random forests, however in order not to advantage our method, prediction errors are calculated with the usual $L^2$ Euclidean distance (time by time) which benefits to the standard approaches like CART trees, RF and FDboost.

\begin{figure}
\centering
\includegraphics[width=\textwidth]{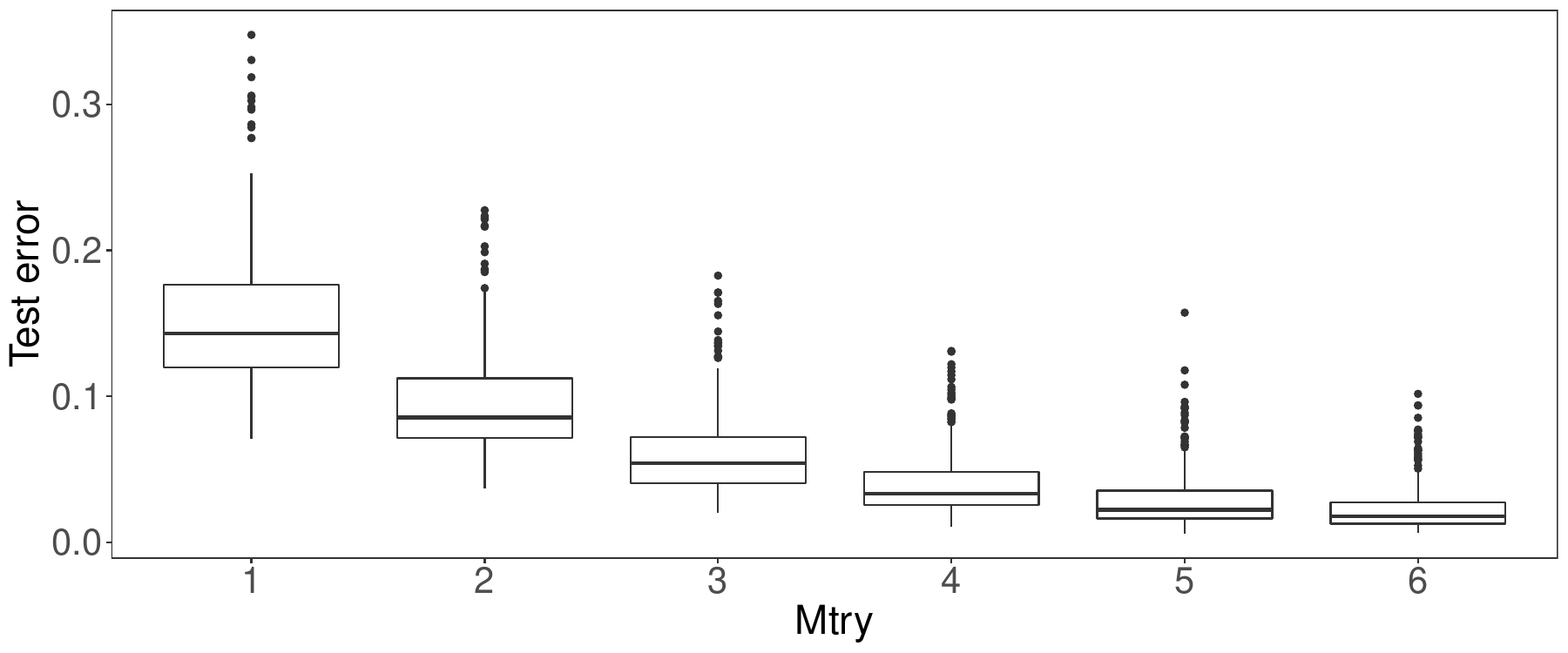}
\caption{Boxplots of the prediction error of the Fréchet random forests method according to the \texttt{mtry} parameter. Prediction errors are calculated on 100 datasets of size $n=100$ simulated according to models \eqref{simX} and \eqref{simY} of the first scenario.}
\label{fig:mtry}
\end{figure}

The number of randomly drawn variables \texttt{mtry} at each node has usually a strong impact on random forests performance: if \texttt{mtry} is too small, individual trees
would give too poor predictions, and if \texttt{mtry} is too high, the
collection of trees could be not diverse enough (\cite{Diaz06};\cite{genuer2008random}). As illustrated in Figure~\ref{fig:mtry} the prediction error (MSE) of the Fréchet random forest decreases as the value of the \texttt{mtry} increases. In all our experiments in the first scenario both in the 2 and 3 temporal behavior functions schemes, we chose \texttt{mtry}=5 and $q=250$ (justified by the fact that, in this experiment, the OOB error stabilizes as soon as $100$ trees are included in the forest). The standard random forest was composed of 500 trees and the \texttt{mtry} parameter was optimized to 2. The number of iterations for FDboost is selected between 1 and 500 through the internal procedure of the package.

\begin{figure*}[!ht]
\begin{center}
\includegraphics[width=\textwidth]{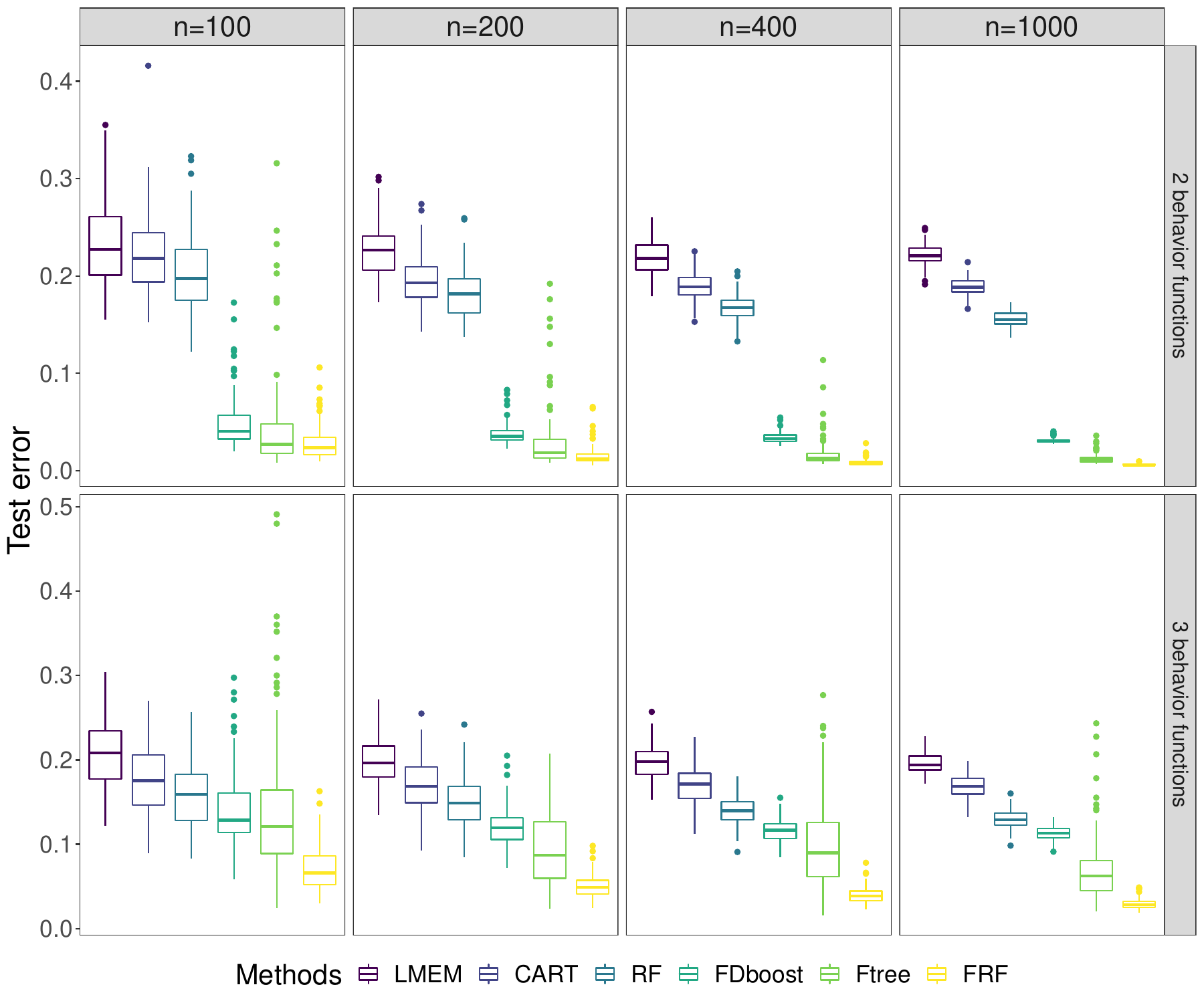}
\caption{Boxplots of the prediction error (MSE) of the Linear mixed effects model (LMEM), CART tree, random forests (RF), FDboost, Fr\'echet tree (Ftree) and Fr\'echet random forest (FRF) methods estimated on 100 datasets simulated according to the simulation scheme of the first scenario for $n=100,\ 200,\ 400$ and $1000$ sample sizes.}
\label{completesim2}
\end{center}
\end{figure*}

\begin{figure*}[!ht]
\begin{center}
\includegraphics[width=0.9\textwidth]{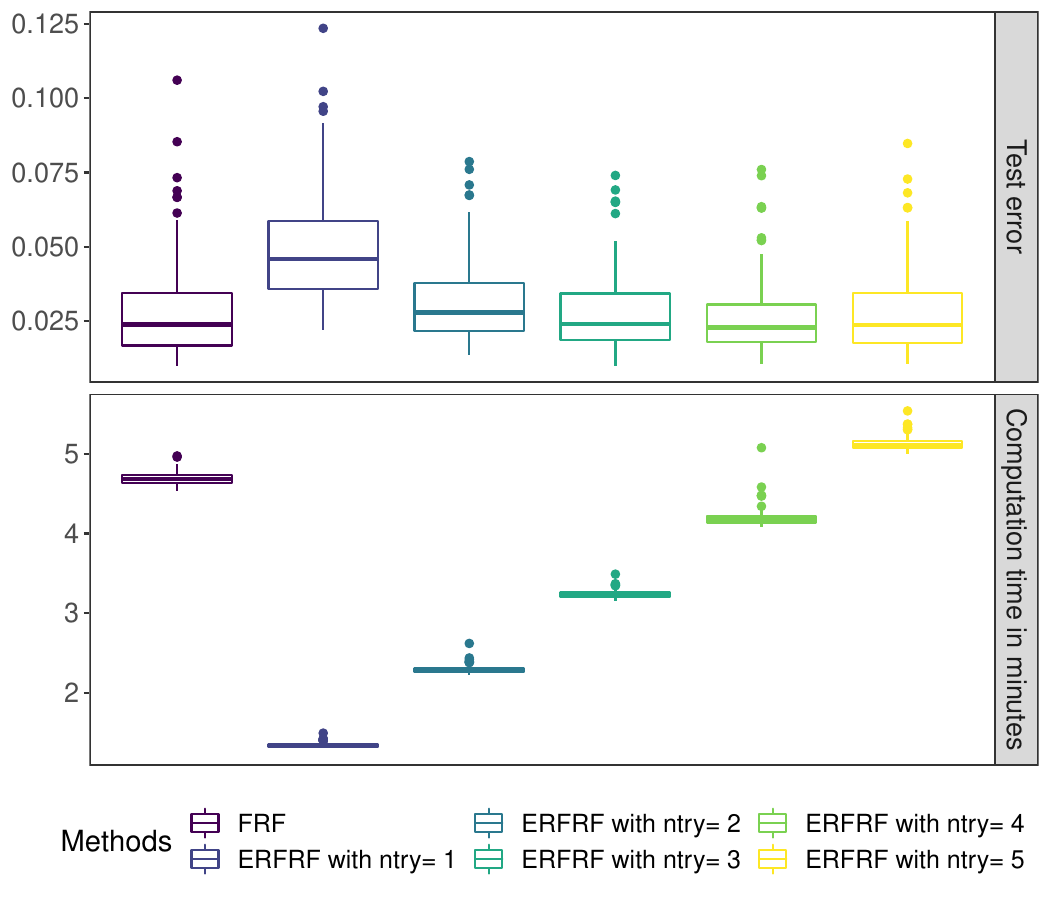}
\caption{Boxplots of the prediction error (MSE) and computation times estimated over 100 datasets of sample size $n=100$ simulated under models \eqref{simX} and \eqref{simY} for Fréchet RF (FRF) method and Extremely Randomized Fréchet RF (ERFRF) method with different values of \texttt{ntry}.}
\label{BoxRandom}
\end{center}
\end{figure*}

\paragraph{Numerical comparison} For any sample size, FDboost, Fr\'echet tree, and Fr\'echet random forests clearly outperform the standard LMEM, CART and RF methods in both the 2 and 3 temporal behaviors schemes. Not surprisingly, the transition from a Fr\'echet tree to a Fr\'echet RF greatly improves predictive capacity by reducing both prediction error and error variance. For instance, in the 2 temporal behavior scheme when $n=100$, the estimated MSE obtained with a Fr\'echet tree is 0.047 while the one obtained with a Fr\'echet RF is 0.028 which is a 40\% decrease in prediction error; this reduction is, for each sample size, always between 40\% and 60\% for the schemes with 2 and 3 temporal behaviors.
Even though FDboost (our principal competitor) shows very good performances, Fr\'echet tree and Fr\'echet RF are the methods that obtain the lowest prediction errors for all sample sizes and schemes. More precisely, for small dataset ($n=100$) in the 2 behavior functions scheme FDboost obtains an estimated MSE of 0.05 while Fr\'echet tree and Fr\'echet RF obtain respectively 0.047 and 0.028 while for large dataset ($n=1000$) the estimated MSE of FDboost is 0.031 and the Fr\'echet tree and Fr\'echet RF estimated MSE are respectively 0.012 and 0.006. Moreover, we can notice that in the scheme with 3 temporal behaviors the prediction error obtained by FDboost is only about 10\% lower than the one obtained by a standard RF, while the MSE obtained by a Fréchet RF is at least 50\% lower than standard RF  (and even keeps a gap of 80\% when $n=1000$). It is worth noting that FDboost obtained an MSE between 75\% and 85\% lower than standard RF in the 2 temporal behavior scheme. We can see here that the change from the scenario with 2 temporal behaviors to 3 temporal behaviors has a greater impact on FDboost than on Fréchet RF.\\
Finally, note that the prediction error of the FDboost, Fr\'echet tree and Fr\'echet RF methods decreases as the sample size $n$ increases which is not the case with other methods that keep a stable prediction error. Additionally, this decrease is much larger with the Fr\'echet tree and Fr\'echet RF methods than with the FDboost method. Moreover, the error prediction of the Fr\'echet RF seems to converge to zero as $n$ tends to infinity in both schemes. In the rest of this section, the objective being to present the advantages in terms of flexibility of the Fr\'echet RF method as well as to compare them to their extremely randomized version, we only consider the scheme with 2 average behavior functions.\\

The extremely randomized version of Fréchet random forests introduced in Section~\ref{ERFRF} has some advantages over the Fréchet RF method. In particular, they are easy to implement, can be used for any type of data and reduce calculation times. In order to verify this claim we calculate the prediction error obtained by extremely randomized Fréchet forests (ERFRF) for different values of \texttt{ntry} on 100 data sets of size $n=100$ simulated according to the first scenario. As shown in Figure~\ref{BoxRandom}, the prediction error of the ERFRF method decreases as the value of the \texttt{ntry} increases. When \texttt{ntry} is large enough (here \texttt{ntry}=3), the error obtained by ERFRF is similar to that obtained by Fréchet RF. Moreover, the execution time of an ERFRF is much lower than that of a Fréchet RF. For example, the build time of an Fréchet RF is 281 seconds while the build time of an ERFRF with ntry=3 is 191 seconds which is 30\% lower. Similar results are obtained on larger datasets (not shown here).\\

\begin{figure*}[!ht]
\begin{center}
\includegraphics[width=\textwidth]{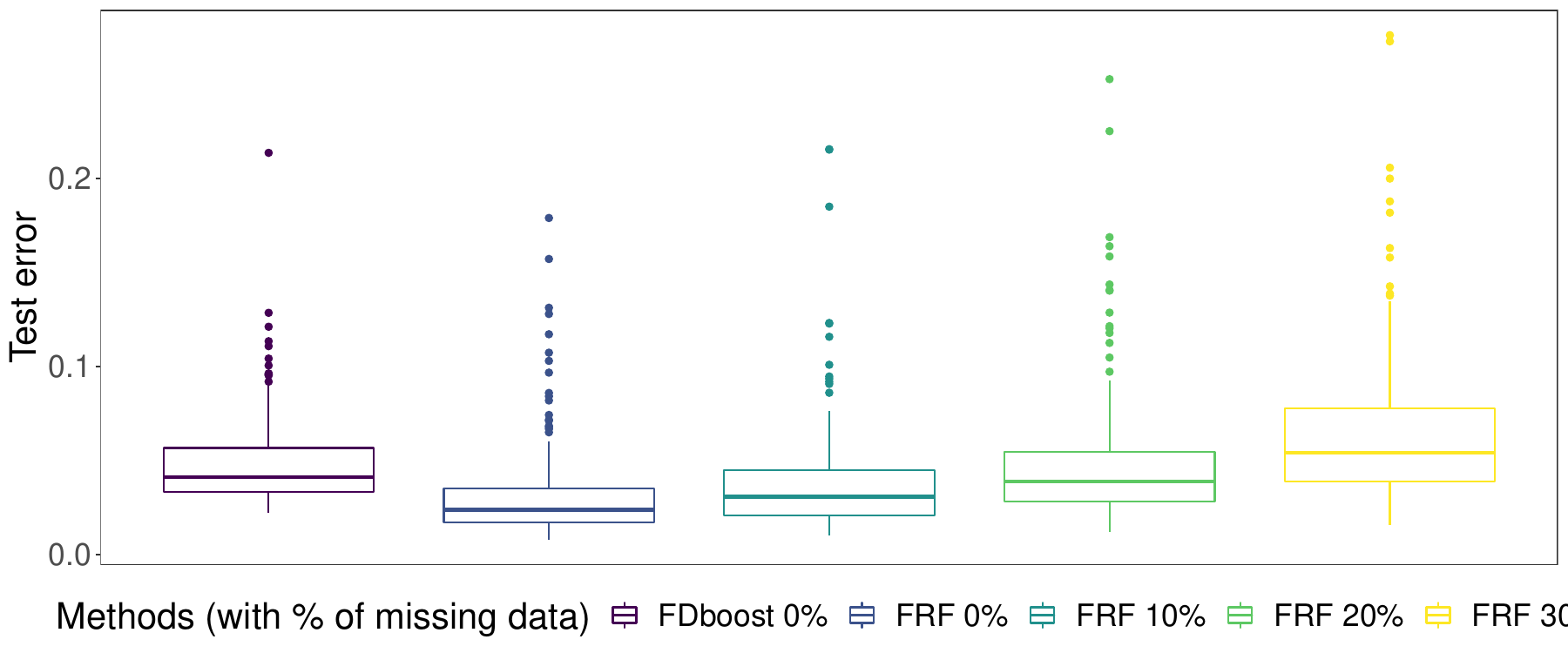}
\caption{Boxplots of the estimated prediction error over 100 datasets of sample size $n$=100 simulated under models~\eqref{simX} and ~\eqref{simY} for FDboost and Fréchet RF (FRF) methods based on the number of missing observations. }
\label{missing}
\end{center}
\end{figure*}

\paragraph{Robustness to missing data} As mentioned in the presentation of the Fréchet distance at the beginning of this section, using the Fréchet distance allows to calculate the distance between two curves measured at different times. Thus, having missing observation times for some curves does not prevent the construction of the trees, as long as not all observation times are missing for a given curve. In order to study the robustness of Fréchet RF to missing observations, we simulate new datasets with $n=100$ individuals according to models~\eqref{simX} and \eqref{simY} by randomly removing 10\%, 20\% and 30\% of the observation times for each curve. It is important to note that the removed observation times are different for each curve. For example, the observations removed for the first variable of the first individual will not necessarily be the same as those removed for the second or third variable or even the output curve of the same individual. It is then impossible to use the standard LMEM, RF and FDboost methods (it is always possible to use the CART method by removing the missing observations for the output curves). As shown in Figure~\ref{missing}, the prediction error obtained by Fréchet RF increases slightly as the percentage of missing observations increases. Moreover, the prediction error obtained with Fréchet RF on simulated data sets with 30\% missing data remains competitive with that obtained by the FDboost method on datasets without missing observations. There are two properties that allow robustness to missing data. The first one comes from the functional data framework and thus to consider that the observations coming from the same individual form a curve.  Indeed, even if some measurement times are missing, as long as there are still some points we still have a curve. The second one comes from the Fréchet metric used which allows us to calculate the distance between curves which are not observed at the same measurement times.

\begin{figure*}[!ht]
\begin{center}
\includegraphics[width=\textwidth]{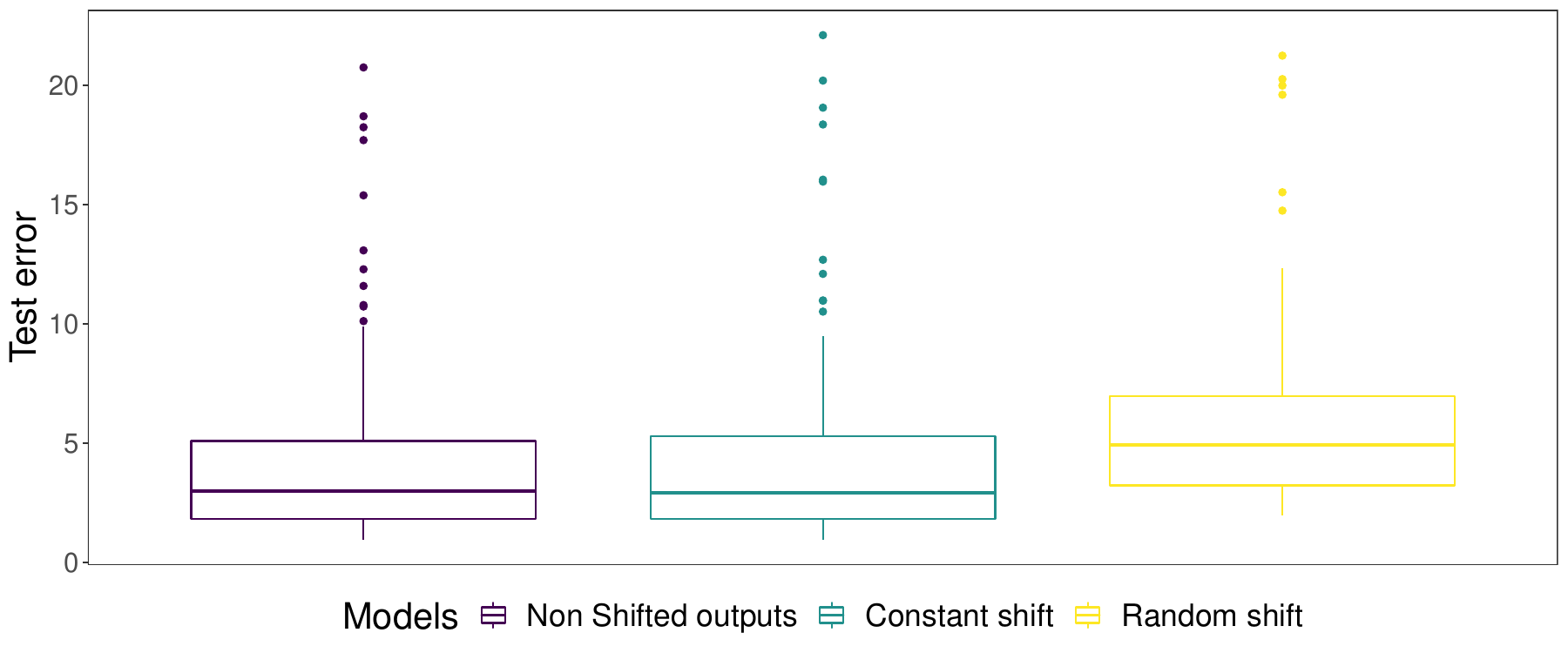}
\caption{ Boxplots of the estimated prediction error over 100 data sets of size $n$=100 simulated under the first scenario for the FRF method based on the time shift applied to the output curves.}
\label{shift}
\end{center}
\end{figure*}

\paragraph{Robustness to time shifts} It is rather common in applications to have a response variable observed after the measurement times of the input variables. In order to study the stability of Fréchet RF method to time shifts, we transform the output curves by shifting them: i) by the same time shift of 1 for all the curves, i.e., the output curves are observed on windows $[1,2]$ instead of $[0,1]$ (keeping the same shapes); ii) by randomly shifting each of them according to a uniform $\mathcal{U}([0,0.5])$, making the windows of observation of the output curves all different in this case (see Figure~\ref{compY} in Appendix~\ref{app:AN2} for the simulated dynamics according to the time shifts). When all the outputs are all translated by a different parameter it is impossible to use the Euclidean distance since the output curves are observed on windows that are not exactly the same. Here we compute the prediction error of the Fréchet RF according to the Fréchet distance (which is the distance used to build the Fréchet RF). As illustrated in Figure~\ref{shift}, the constant time shift for the response curves has no influence on the Fréchet RF prediction error. When the offsets are randomly drawn for each output curve, the prediction error increases slightly to an average error of 5.2. As an example, the prediction error of FDboost computed with the Fréchet distance on simulated data without time shifts is 11.4. We refer to \cite{10.1371/journal.pone.0150738} for a complete presentation of the Fréchet averaging algorithm for curves, even when time shifted.

\begin{figure*}[!ht]
\begin{center}
\includegraphics[width=\textwidth]{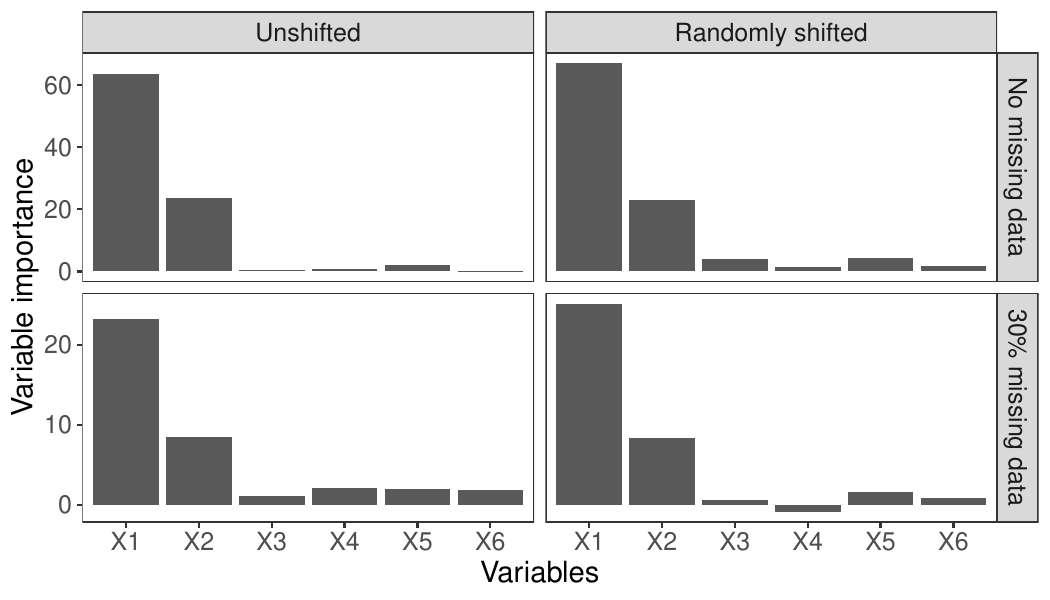}
\caption{Barplots of the Fréchet RF variable importance scores, obtained on 4 datasets simulated according to model~\eqref{simX} and model~\eqref{simY}. The results in the left-hand column are obtained on the simulated datasets without time shift while the right-hand column contains those obtained with a random time shift on the output curves. The results on the first row are those obtained on the simulated data sets without missing data while those on the second row are those obtained on the simulated data sets with 30\% missing data on the input and output curves.}
\label{VIcurves}
\end{center}
\end{figure*}

\paragraph{Variables importance sensitivity} Finally, Figure~\ref{VIcurves} gives the importance scores of variables calculated with the Fréchet RF method on 4 datasets of size $n=100$ simulated according to models~\eqref{simX} and~\eqref{simY}:\begin{enumerate}
\item With no time shifts on the output curves and no missing observation times.
\item With random time shifts according to a uniform $\mathcal{U}([0,0.5])$ on the output curves but with no missing measurement times.
\item With no time shifts but with 30\% missing observation times.
\item With 30\% missing observations and time shifts on the output curves.
\end{enumerate}
This graph shows that neither time shifts nor missing observation times have an impact on the importance of the variables. Indeed, the first two variables (those related to the output variable) are always the ones with the highest importance scores. The other four variables (unrelated to the output variable) have extremely low importance scores compared to the first two variables. 

As a conclusion, we illustrate the superiority on longitudinal data (in terms of prediction error) of the Fréchet trees and Fréchet RF methods compared to the standard LMEM, CART, RF methods as well as the longitudinal boosting method FDboost. In addition, we illustrate the great robustness of the method to missing data and time shifts, both in terms of prediction error and the importance of the variables. 
Lastly, we show that the extremely randomized variant ERFRF can obtain a prediction error similar to that of Fréchet RF while having lower computation times, making it a method of choice for analyzing very large datasets.  

\subsubsection{Second scenario}
\label{sec:res2}

The Fréchet distance is used on curve spaces while the standard Euclidean distance is used on scalar spaces and image variables. Since there is no comparison with other methods in this scenario, the OOB error will be used as a measure of the performance of the Fréchet RF. Throughout this section we study the ERFRF method, the version implemented in our package \texttt{FrechForest} that can handle curves, images and scalars as inputs.\\

\begin{figure}[!ht]
\centering
\begin{tikzpicture}
  \node[draw,circle] (C1) at (-2, 0) {$t_1$};
  \node[draw,circle] (C2) at (-3.75, -2) {$t_2$};
  \node[draw,circle] (C3) at (-0.25, -2) {$t_3$};
  \node[draw,circle] (C4) at (-4.75, -4) {$t_4$};
  \node[draw,circle] (C5) at (-2.75, -4) {$t_5$};
  \node[draw,circle] (C6) at (0.75, -4) {$t_7$};
  \node[draw,circle] (C7) at (-1.25, -4) {$t_6$};
  \draw (C1) -- (C2);
  \draw (C1) -- (C3);
  \draw (C2) -- (C4);
  \draw (C2) -- (C5);
  \draw (C3) -- (C6);
  \draw (C3) -- (C7);
  \node (s1) at (-2, -1) {\scriptsize $S^{(1)}\leq 1.11$};
  \node (s2l) at (-0.25, -3) {\scriptsize $S^{(1)} \leq 1.67$};
  \node (s3) at (-3.75, -3) {\scriptsize $S^{(1)}\leq 0.53$};
  \node (name) at (0,1) {\textbf{1)}};

  \node[draw,circle] (cC1) at (5, 0) {$t_1$};
  \node[draw,circle] (cC2) at (3.25, -2) {$t_2$};
  \node (cX2) at (3.35,-0.75) {\includegraphics[width=4ex]{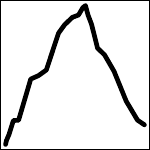}};
  \node[draw,circle] (cC3) at (6.75, -2) {$t_3$};
  \node (cX3) at (6.65,-0.75) {\includegraphics[width=4ex]{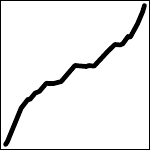}};
  \node[draw,circle] (cC4) at (2.25, -4) {$t_4$};
  \node (cX121) at (2.25,-2.75) {\includegraphics[width=4ex]{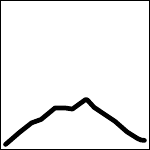}};
  \node[draw,circle] (cC5) at (4.25, -4) {$t_5$};
  \node (cX122) at (4.25,-2.75) {\includegraphics[width=4ex]{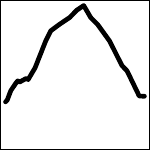}};
  \node[draw,circle] (cC6) at (7.75, -4) {$t_7$};
  \node (cX122) at (7.75,-2.75) {\includegraphics[width=4ex]{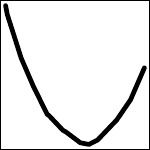}};
  \node[draw,circle] (cC7) at (5.75, -4) {$t_6$};
  \node (cX122) at (5.75,-2.75) {\includegraphics[width=4ex]{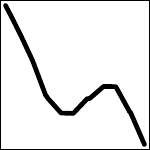}};
  \draw (cC1) -- (cC2);
  \draw (cC1) -- (cC3);
  \draw (cC2) -- (cC4);
  \draw (cC2) -- (cC5);
  \draw (cC3) -- (cC6);
  \draw (cC3) -- (cC7);
  \node (cs1) at (5, -1) {\scriptsize $X^{(1)}$};
  \node (cs2l) at (6.75, -3) {\scriptsize $X^{(2)}$};
  \node (cs3) at (3.25, -3) {\scriptsize $X^{(1)}$};
  \node (cname) at (7,1) {\textbf{2)}};

  \node[draw,circle] (C1) at (-2, -6) {$t_1$};
  \node[draw,circle] (C2) at (-3.75, -8) {$t_2$};
  \node (X2) at (-3.65,-6.75) {\includegraphics[width=4ex]{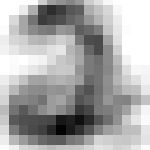}};
  \node[draw,circle] (C3) at (-0.25, -8) {$t_3$};
  \node (X3) at (-0.35,-6.75) {\includegraphics[width=4ex]{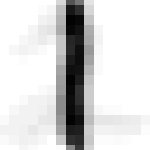}};
  \node[draw,circle] (C4) at (-4.75, -10) {$t_4$};
  \node (X121) at (-4.75,-8.75) {\includegraphics[width=4ex]{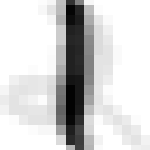}};
  \node[draw,circle] (C5) at (-2.75, -10) {$t_5$};
  \node (X122) at (-2.75,-8.75) {\includegraphics[width=4ex]{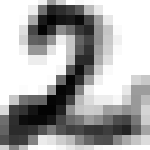}};
  \node[draw,circle] (C6) at (0.75, -10) {$t_7$};
  \node[draw,circle] (C7) at (-1.25, -10) {$t_6$};
  \draw (C1) -- (C2);
  \draw (C1) -- (C3);
  \draw (C2) -- (C4);
  \draw (C2) -- (C5);
  \draw (C3) -- (C6);
  \draw (C3) -- (C7);
  \node (s1) at (-2, -7) {\scriptsize $I^{(2)}$};
  \node (s2l) at (-0.25, -9) {\scriptsize $S^{(1)} \leq 0.66$};
  \node (s3) at (-3.75, -9) {\scriptsize $I^{(1)}$};
  \node (name) at (0,-5) {\textbf{3)}};

  \node[draw,circle] (C1) at (5, -6) {$t_1$};
  \node[draw,circle] (C2) at (3.25, -8) {$t_2$};
  \node (X2) at (3.35,-6.75) {\includegraphics[width=4ex]{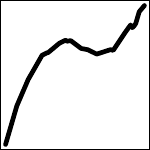}};
  \node[draw,circle] (C3) at (6.75, -8) {$t_3$};
  \node (X3) at (6.65,-6.75) {\includegraphics[width=4ex]{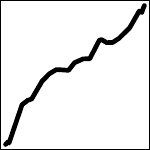}};
  \node[draw,circle] (C4) at (2.25, -10) {$t_4$};
  \node (X121) at (2.25,-8.75) {\includegraphics[width=4ex]{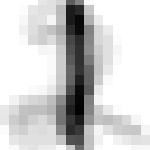}};
  \node[draw,circle] (C5) at (4.25, -10) {$t_5$};
  \node (X122) at (4.25,-8.75) {\includegraphics[width=4ex]{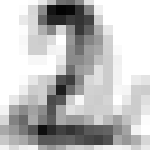}};
  \node[draw,circle] (C6) at (7.75, -10) {$t_7$};
  \node[draw,circle] (C7) at (5.75, -10) {$t_6$};
  \draw (C1) -- (C2);
  \draw (C1) -- (C3);
  \draw (C2) -- (C4);
  \draw (C2) -- (C5);
  \draw (C3) -- (C6);
  \draw (C3) -- (C7);
  \node (s1) at (5, -7) {\scriptsize $X^{(1)}$};
  \node (s2l) at (6.75, -9) {\scriptsize $R^{(1)} \leq 0.56$};
  \node (s3) at (3.25, -9) {\scriptsize $I^{(1)}$};
  \node (name) at (7,-5) {\textbf{4)}};

 \end{tikzpicture}
\caption{Examples of 4 extremely randomized trees of depth 2 built on $n=100$ simulated observations according to the second scenario. The 4 trees are constructed from the input variables of: 1) scalars only; 2) curves only; 3) images and scalars; 4) curves, images and scalars. Below each node is indicated the split variable. To the left and right of each node are indicated the representative elements of the right and left child nodes for the split variable in question. For example for model 3) the split variable of the root node is $I^{(2)}$, the images of the variable $i^{(2)}$ which are closer to the image on the left (for the Euclidean distance), a blurred 2, go into the node $t_2$ while those closer to 1 go into the node $t_3$.}
\label{fig:trees}
\end{figure}
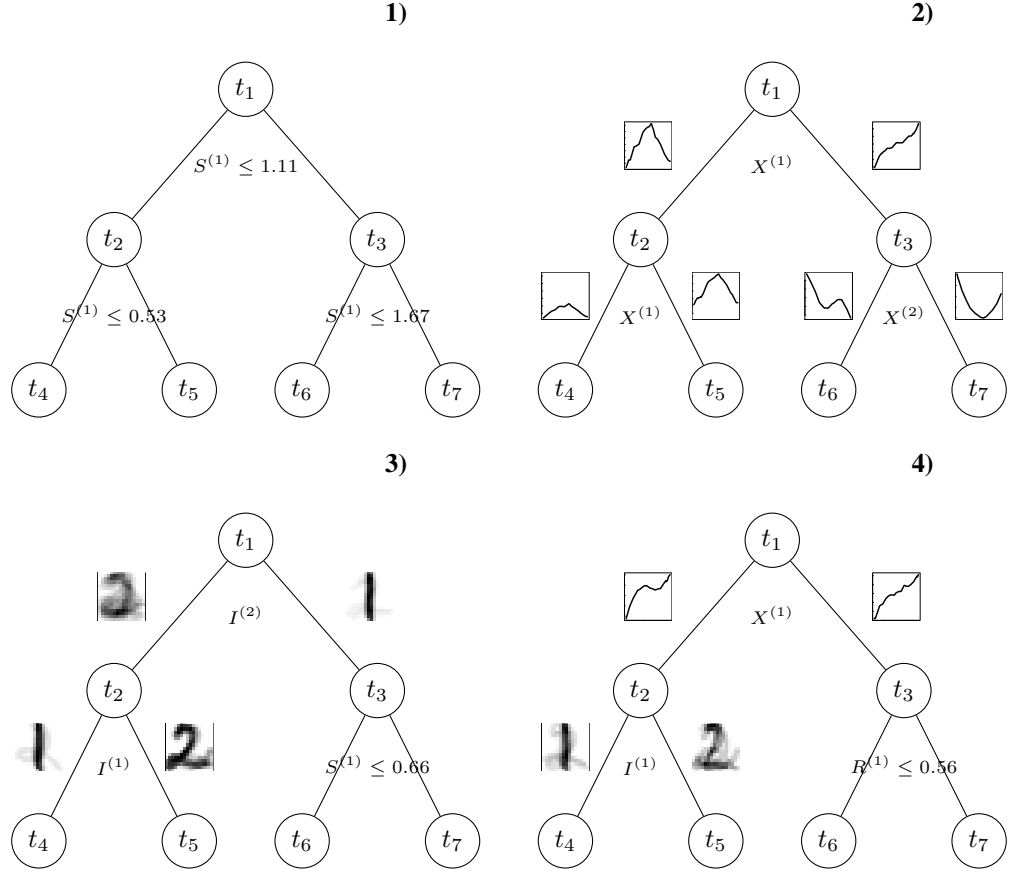

We study the OOB error obtained by ERFRF according to the types of input variables (images, curves or scalars) on 100 datasets of size $n=$100 simulated according to the second scenario. We consider the following models: \begin{enumerate}
\item\label{mod:scalar} Only scalar variables $R^{(1)}$ and $R^{(2)}$ are used to predict output curves.
\item\label{mod:curves} Only curve variables $X^{(1)},\ldots,X^{(6)}$ are used to predict output curves.
\item\label{mod:imsca} Image variables $I^{(1)}$ and $I^{(2)}$ and scalar variables are used.
\item\label{mod:full} all variables \emph{i.e. }curves, scalars and images are used to predict output curves. 
\end{enumerate}
Note that case~\ref{mod:curves} corresponds to the first simulation scenario. Figure~\ref{fig:trees} shows an example of an extremely randomized Fréchet tree of depth 2 (only the first three splits are shown here) for each model above. When the models incorporate different types of inputs, in the case of models 3) and 4), the constructed trees are mixed in the sense that they can alternate the split spaces. For example, in the case of model 4), Figure 9 shows an example of a tree with the first three splits in the three different types of input spaces: curves, scalars and images. 
We chose the parameters \texttt{mtry}=5, $q=250$ and \texttt{ntry}=5 for each model.  

\begin{figure}[!ht]
\centering
\includegraphics[width=\textwidth]{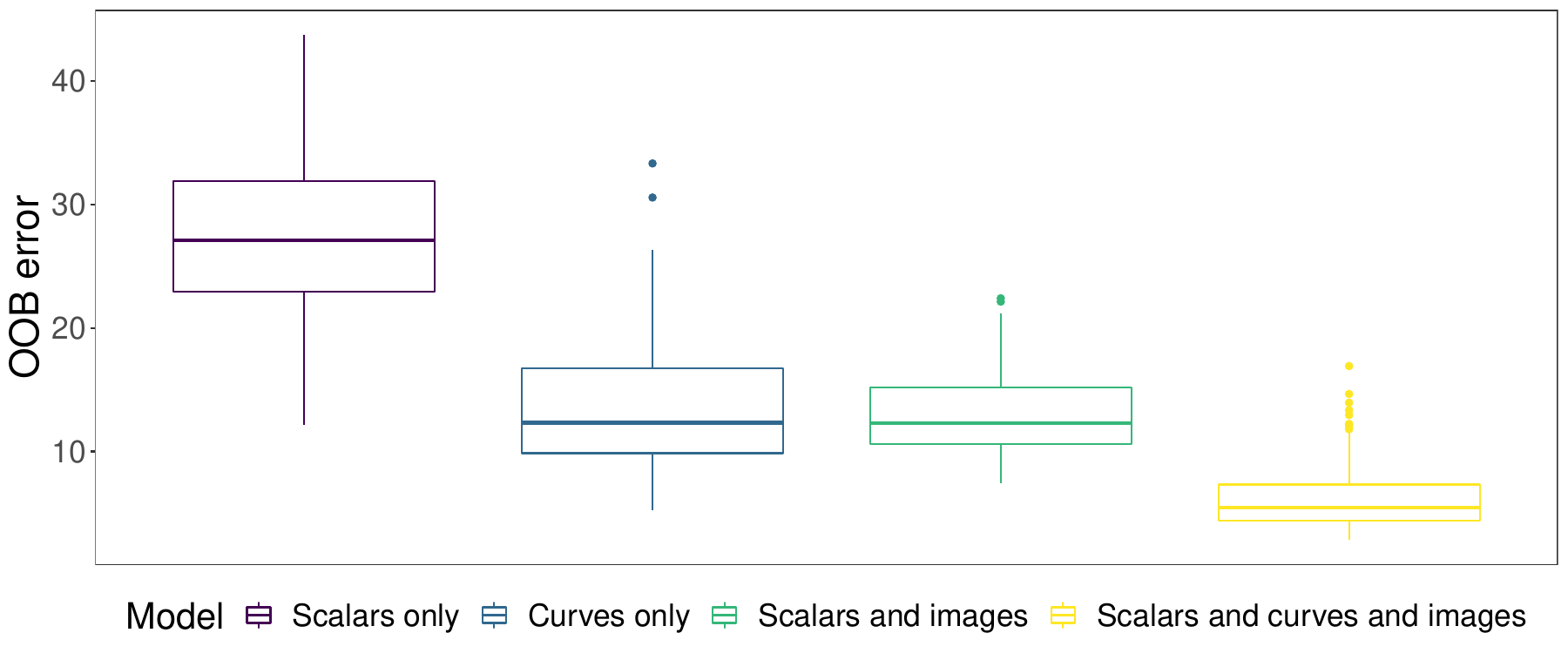}
\caption{OOB errors of the ERFRF method according to the types of input variables. The OOB errors are obtained on 100 data sets of size $n=100$ simulated according to the second scenario.}
\label{boxsce2}
\end{figure}

As shown in Figure~\ref{boxsce2}, the highest OOB error is obtained when only scalar variables are used. When the image variables are added to the scalar variables, the OOB error is the same as the one obtained on the model using only the input curves. This was expected since the input curve variables provide the same information as the image and scalar variables combined. More precisely, the input curve variables provide both information on the shape of the output curves as well as on their amplitude, whereas the information on the shape is only provided by the images and the ones on the amplitude is only provided by the scalars. Individually, the input variables of images or scalars provide only part of the information that is provided by the input variables of curves. Finally, when the image and scalar variables are added to the curve variables, the OOB error of the ERFRF decreases. This is explained by the fact that in some cases, when the contraction or dilation of the input curves is too large, the dilated or contracted curves may have a very different shape than their initial shape and thus lose the information they brought due to their shape. Thus the addition of image variables allows to always have access to information on the shapes of the output curves. Finally, the results of these simulations emphasizes the main strength of the ERFRF method, which is to handle heterogeneous data, i.e. input and output variables of different natures. 

\section{Application to air quality prediction}
\label{sec:app}

The \texttt{airquality} dataset\footnote{The data are publicly available at \url{https://archive.ics.uci.edu/dataset/360/air+quality}} \citep{Devito2008} contains $9358$
observations of hourly averaged atmospheric pollutants concentrations both from
sensors and from a certified analyzer.
The observations correspond to daily measurements from March 2004 to
February 2005, on the field in a significantly polluted area, at road level,
within an Italian city. In our analysis, we only consider ground truth
measurements (given by the certified analyzer) and we take the carbon monoxide
(CO) concentration as the response (as in \cite{Luo2023}).
The other measured pollutants concentrations are nitrogen dioxide (NO$_2$),
total nitrogen oxides (NO$_x$) and benzene(C$_6$H$_6$). Note that we did not
consider the non-methane hydrocarbons (NMHC) pollutant because it presents more
than $90\%$ missing data. In addition, the hourly average temperature (Temp),
relative (RH) and absolute humidity (AH) are also reported.

Our aim is, for a given day, to predict the CO concentration curve corresponding
to the second-half of the day (from 12 a.m. to 23 p.m) using the other variables
curves restricted to the first-half of the day (from 0 a.m. to 11 a.m.). Hence,
we consider the statistical unit as the day of the year (ending up with $304$
units after removing the days with too many missing values), and for each unit
we have $6$ input curve variables and $1$ output curve variable. We then have a
function-to-function prediction problem for which we use Fréchet random forests
(FRF).

We also apply standard random forests (RF) to the same data but now
with hours as statistical units : for each hour (of all day of the year) we have
$6$ input scalar variables and $1$ scalar output variable. In this case, the
output variable is the CO concentration measured $12$ hours later than the time
of measurement of the input variables (we then get $3648 = 304 \times 12$ units
in this case). The objective of this second analysis is to study if the fact
that FRF take into account the curve structure of the data helps to get better
predictions or not, compared to RF that ignore that structure (and thus consider
all observations as independent of each other).

To compare the prediction performance of both approaches, we use the mean
squared error at the hour level:
$\frac{1}{N} \sum_{i=1}^N (\widehat{h}(x_i) - y_i)^2$, where $N$ is the total
number of hours of the year, $x_i$ and $y_i$ are the input variables and the
output variable resp., observed at hour $i$, and $\widehat{h}$ denotes either
the FRF or the RF predictor. This is directly calculable for RF using OOB
predictions. However, it has to be recomputed for FRF since their OOB
predictions have a curve structure (see Section~\ref{sec:oob}). This is done by
computing the squared difference between predictions and actual CO
concentrations pointwise for those predicted curves.

We stress that, as in previous sections, we keep the Fréchet distance for the
space of curve variables. Hence, the FRF method is not parameterized to
optimize the pointwise error of its predictions. The \texttt{mtry} parameter was
optimized for both methods leading to \texttt{mtry} $=3$ for RF and
\texttt{mtry} $=2$ for FRF, while $500$ trees were built in each
forest  \footnote{The R code of this analysis is available at \url{https://github.com/sistm/airquality_FrechForest}}.

We ran the two methods 20 times on the same data and computed the median and the
interquartile range (IQR) on the 20 obtained errors. The RF predictor reached a
median error of $1.486$ (IQR: $1.482 - 1.488$), while FRF managed to get
$1.160$ (IQR: $1.155 - 1.173$), which corresponds to an error reduction of
$22\%$. Interestingly, the variable importance scores
(see Figure~\ref{fig:varImpComp} in Appendix~\ref{ANairqual})
were also quite different with the two methods:
while NO$_x$ and NO2 variables were the most important
variables, and C6H6 the least one, for RF; the two most important variables for
FRF were NO$_x$ and C6H6. In other words, C6H6 did not seem to help in the
independent hourly data case, while it was among the two most useful variables
to predict the output in the daily curve data case.

\section{Discussion}
\label{sec:discussion}

Two new tree-based methods, Fr\'echet trees and Fr\'echet random forests, for general
metric spaces-valued data were introduced.
Let us emphasize that the proposed methods are very general. Indeed, input variables can thus all be of different kinds, each one having its own metric, and the kind of the output variable can also be a different one.

The example of learning curve shapes in the context of longitudinal/functional data was presented to illustrate the capacity of the methods to learn from data in unordered metric spaces. A simulation study in this framework demonstrated the superiority of Fr\'echet trees and forests over the existing classical methods,  both in terms of prediction error as well as robustness and flexibility. An important aspect highlighted in our study is the great robustness of Fréchet trees and Fréchet random forests. Indeed, our simulations illustrated the ability to handle missing data as well as different observation times for the different variables, which is common in longitudinal datasets. Two other simulation scenarios demonstrated the capacity of the methods to simultaneously handle data of different natures such as curves, images, scalars, factors, shapes, etc. This great flexibility allows the construction of more efficient predictors while being able to compare the information provided by each of these variables of different natures thanks to the importance score. Finally, within the framework of a study on air quality, we highlighted the superiority of the Fréchet RF method over standard RF. We illustrated that regression on curve shapes could greatly improve the prediction error while using different information from input variables.

However, there are two main limitations to Fr\'echet trees and forests: the first one
is that the Fr\'echet mean has to exist in the output space \citep{legouic:hal-01163262} and has to be fairly approximated.
The second concerns the computation time. Indeed, as mentioned in the section \ref{res1}, in the implementation of our R package \texttt{FrechForest} the emphasis has largely been put on flexibility, which allows to analyze a very large spectrum of data such as images, curves, scalars, factors and shapes. However, this implementation has not been optimized for optimal computation times and then Fr\'echet random forests can still be computationally intensive. This problem can be alleviated by the fact that, as all forests methods, they are
easily parallelized (the different trees can be built in parallel).

For the theoretical side, we have proved a consistency result for purely uniformly random trees in the case where the input space is $[0,1]$ and the output space is a general metric space. Obviously, it would be interesting to consider trees in which the splitting criterion is the 2-means function and manage to prove properties as in Theorem~\ref{T1}. However, this is a quite complex problem which is out of the scope of this paper. For the practical side, we are developing a new implementation of Fréchet random trees and forests in Julia language. The package currently under development is called \texttt{ExtraFrech} and focuses on performance and the ability to analyze large data sets. In the current version, preliminary tests show that our new implementation is competitive with the R package \texttt{randomForest}.  We are also working on an efficient implementation of metrics adapted to image data, such as the Wassertein distance \citep{doi:10.1137/1118101}, in order to apply the Fréchet RF method to large brain imaging databases. 

\appendix
\section{Proof of Theorem~\ref{T1}}
\label{AN1}
First, we demonstrate the point-wise consistency given by \eqref{eq:conv}. We introduce the following quantity \begin{equation}
r_n\left(x,y\right)=\frac{\frac{1}{n}\sum_{i=1}^nd^2(Y_i,y)\textbf{1}\{X_i\in \pi_n[x]\}}{\mathbb{P}\left(X\in \pi_n[x]\right)}
\end{equation}
From \ref{eq:est} we have $T_n(x)=\underset{y\in\mathcal{Y}}\argmin\ r_n\left(x,y\right)$. First, we use the following classical upper bound in $M$-estimation:

\begin{align}
\label{ineq:sup}
r\left(x, T_n(x)\right)-\underset{y\in\mathcal{Y}}\min\ r\left(x,y\right)&= r\left(x, T_n(x)\right)-r_n\left(x,T_n(x)\right)+r_n\left(x,T_n(x)\right)-\underset{y\in\mathcal{Y}}\min\ r\left(x,y\right)\notag\\
&= r\left(x, T_n(x)\right)-r_n\left(x,T_n(x)\right)+r_n\left(x,T_n(x)\right)- r\left(x,\phi^{*}(x)\right)\notag\\
&\leq r\left(x, T_n(x)\right)-r_n\left(x,T_n(x)\right)+r_n\left(x,\phi^{*}(x)\right)- r\left(x,\phi^{*}(x)\right)\notag\\
&\leq 2\underset{y\in\mathcal{Y}}\sup\left|r_n\left(x,y\right)-r(x,y)\right|
\end{align}
We are going to decompose the above supremum in several terms that we are going to appropriately upperbound to obtain their decay to zero under the assumptions Theorem \ref{T1}. Consider a $\delta$ covering of $\mathcal{Y}$ with centers $\{y_{\alpha}\}_{\alpha=1}^Q$ where $Q = N(\delta,\mathcal{Y},d)$. Thus, for every $y\in\mathcal{Y}$, there is $\alpha = \alpha_y \in\{1,\ldots,Q\}$ such as $d\left(y,y_{\alpha}\right)<\delta$. We introduce the following quantity \begin{equation}
r^{E}\left(x,y\right)=\frac{\mathbb{E}\left(d^2(Y,y)\textbf{1}\{X\in \pi_n[x]\}\right)}{\mathbb{P}\left(X\in \pi_n[x]\right)}
\end{equation}
Then, the following decomposition is used
\begin{multline}
\label{eq:decomp}
r_n\left(x,y\right)-r(x,y)=\underbrace{r_n\left(x,y\right)-r_n\left(x,y_{\alpha}\right)}_{(i)} + \underbrace{r_n\left(x,y_{\alpha}\right) - r^{E}\left(x,y_{\alpha}\right)}_{(ii)}\\ +\underbrace{r^{E}\left(x,y_{\alpha}\right) - r^{E}\left(x,y\right)}_{(iii)} + \underbrace{r^{E}\left(x,y\right) - r\left(x,y\right)}_{(iv)}
\end{multline}
We are now going to derive upper bounds for each of the four terms above  that do not depend on $y$.  Let us start with the term $(i)$ of \eqref{eq:decomp}, we introduce the following event
$$
\mathcal{E}_n = \left\{ \left|\frac{\frac{1}{n}\sum_{i=1}^{n}\textbf{1}\{X_i\in \pi_n[x]\}}{\mathbb{P}\left(X\in \pi_n[x]\right)}-1\right|<\frac{1}{2} \right\}.
$$
We can upper bound the probability of the complementary of the event  $\mathcal{E}_n$ (denoted $\mathcal{E}_n^c$) as 

\begin{align}
\mathbb{P}\left(\mathcal{E}_n^c\right)& =\mathbb{P}\left(
\left|\frac{\frac{1}{n}\sum_{i=1}^{n}\textbf{1}\{X_i\in \pi_n[x]\}}{\mathbb{P}\left(X\in \pi_n[x]\right)}-1\right|>\frac{1}{2}\right)\\ &\leq \mathbb{P}\left(
\underset{\pi\in\Pi_n}\sup\underset{A\in\mathcal{\pi}}\sum\left|\frac{\frac{1}{n}\sum_{i=1}^{n}\textbf{1}\{X_i\in A\}}{\mathbb{P}\left(X\in A\right)}-1\right|>\frac{1}{2}\right)
\end{align}
Then, we upper bound the last probability using Lemma~\ref{lem2}
\begin{equation}
\label{Nobel}
\mathbb{P}\left(
\underset{\pi\in\Pi_n}\sup\underset{A\in\mathcal{\pi}}\sum\left|\frac{\frac{1}{n}\sum_{i=1}^{n}\textbf{1}\{X_i\in A\}}{\mathbb{P}\left(X\in A\right)}-1\right|>\frac{1}{2}\right)\leq 4\Delta_n^{*}(\Pi_n)2^{\mathcal{C}(\Pi_n)}\exp - \frac{n}{128}
\end{equation}
On the event $\mathcal{E}_n$, one has that $\left|\frac{\frac{1}{n}\sum_{i=1}^{n}\textbf{1}\{X_i\in \pi_n[x]\}}{\mathbb{P}\left(X\in \pi_n[x]\right)}\right|<\frac{3}{2}$ which implies that

\begin{multline}
\label{ineq:diam1}
\left|r_n\left(x,y\right)-r_n\left(x,y_{\alpha}\right)\right|=\left|\frac{\frac{1}{n}\sum_{i=1}^{n}\left(d^2(Y_i,y)-d^2(Y_i,y_{\alpha})\right)\textbf{1}\{X_i\in \pi_n[x]\}}{\mathbb{P}\left(X\in \pi_n[x]\right)}\right|\\=\left|\frac{\frac{1}{n}\sum_{i=1}^{n}\left(d(Y_i,y)-d(Y_i,y_{\alpha})\right)\left(d(Y_i,y)+d(Y_i,y_{\alpha})\right)\textbf{1}\{X_i\in \pi_n[x]\}}{\mathbb{P}\left(X\in \pi_n[x]\right)}\right|\\ \leq 2\diam\left(\mathcal{Y}\right)d\left(y,y_{\alpha}\right)\left|\frac{\frac{1}{n}\sum_{i=1}^{n}\textbf{1}\{X_i\in \pi_n[x]\}}{\mathbb{P}\left(X\in \pi_n[x]\right)}\right|\leq 3\diam\left(\mathcal{Y}\right)\delta
\end{multline}
On the complementary of the event $\mathcal{E}_n$, we use similar arguments and the upper bound $\left|\frac{\frac{1}{n}\sum_{i=1}^{n}\textbf{1}\{X_i\in \pi_n[x]\}}{\mathbb{P}\left(X\in \pi_n[x]\right)}\right| \leq \frac{1}{\mathbb{P}\left(X\in \pi_n[x]\right)}$ to derive that
$$
\left|r_n\left(x,y\right)-r_n\left(x,y_{\alpha}\right)\right| \leq 2\diam\left(\mathcal{Y}\right) \delta \frac{1}{\mathbb{P}\left(X\in \pi_n[x]\right)}.
$$
Therefore, we finally obtain that
$$
\left|r_n\left(x,y\right)-r_n\left(x,y_{\alpha}\right)\right| \leq \diam\left(\mathcal{Y}\right)\delta\left( 3 \mathbb{P}\left(\mathcal{E}_n \right)+ 2 \frac{1 - \mathbb{P}\left(\mathcal{E}_n \right)}{\mathbb{P}\left(X\in \pi_n[x]\right)}\right)
$$
Now we consider the term $(ii)$ in \eqref{eq:decomp}. To this end, we propose to bound the following probability
\begin{align}
\label{eq:all}
\mathbb{P}\left(\underset{\alpha=1,\ldots,Q}\max\ \left|r_n\left(x,y_{\alpha}\right)-r^{E}\left(x,y_{\alpha}\right)\right|>\epsilon\right)&= \mathbb{P}\left(\cup_{\alpha=1}^Q\left\{ \left|r_n\left(x,y_{\alpha}\right)-r^{E}\left(x,y_{\alpha}\right)\right|>\epsilon\right\}\right)\notag\\ & \leq \sum_{\alpha=1}^Q\mathbb{P}\left( \left|r_n\left(x,y_{\alpha}\right)-r^{E}\left(x,y_{\alpha}\right)\right|>\epsilon\right)
\end{align}
For a fixed $\alpha\in\{1,\ldots,Q\}$, we define
$W_i=\frac{d^2(Y_i,y_{\alpha})\textbf{1}\{X_i\in \pi_n[x]\}}{\mathbb{P}\left(X\in \pi_n[x]\right)}$, and we thus have that $$r_n\left(x,y_{\alpha}\right)-r^{E}\left(x,y_{\alpha}\right)=\frac{1}{n}\sum_{i=1}^nW_i-\mathbb{E}\left(W_i\right),$$
which can be controlled thanks to Bernstein's inequality by finding upper bounds on $|W_i|$ and $\var (W_i)$. To this end, we first derive a lower bound on $\mathbb{P}\left(X\in \pi_n[x]\right)$.
\begin{align*}
\mathbb{P}\left(X\in \pi_n[x]\right)&=\mathbb{E}\left(\mathbb{P}\left(X\in \pi_n[x]\ | \mathcal{L}_n\right)\right)\\&=\mathbb{E}\left(\int_{\pi_n[x]} \rho_X(t) dt\right)\geq \rho_{\min}\mathbb{E}\left(\Vol (\pi_n[x])\right)\\&=\rho_{\min}\mathcal{V}_n[x]\qquad\mbox{with}\quad\mathcal{V}_n[x]=\mathbb{E}\left(\Vol (\pi_n[x])\right).
\end{align*}
Therefore, we obtain that $$|W_i|\leq \frac{d^2(Y_i,y_{\alpha})\textbf{1}\{X_i\in \pi_n[x]\}}{\rho_{\min}\mathcal{V}_n[x]}\leq \frac{\diam ^2 \left(\mathcal{Y}\right)}{\rho_{\min}\mathcal{V}_n[x]}$$
Moreover,
\begin{equation*}
\var(W_i)\leq\mathbb{E}\left(W_i^2\right) =\frac{\mathbb{E}\left(d^4(Y_i,y_{\alpha})\textbf{1}^2\{X_i\in \pi_n[x]\}\right)}{\mathbb{P}\left(X\in \pi_n[x]\right)^2}\leq \frac{\diam ^4(\mathcal{Y})\mathbb{P}(X\in \pi_n[x])}{\mathbb{P}(X\in \pi_n[x])^2}\\ \leq\frac{\diam ^4(\mathcal{Y})}{\rho_{\min}\mathcal{V}_n[x]} 
\end{equation*}
Then, by Bernstein's inequality, we have for every $\alpha\in\{1,\ldots,Q\}$ \begin{align*}
\mathbb{P}\left(\left|r_n(x,y_{\alpha})-r^{E}(x,y_{\alpha})\right|>\epsilon\right)&\leq 2\exp\left(\frac{-n\epsilon^2}{\frac{2\diam ^4(\mathcal{Y})}{\rho_{\min}\mathcal{V}_n[x]}+ \frac{2\diam ^2(\mathcal{Y})\epsilon}{\rho_{\min}\mathcal{V}_n[x]}}\right)\\ &=2\exp\left(\frac{-n\epsilon^2\rho_{\min}\mathcal{V}_n[x]}{2\diam ^2(\mathcal{Y})(\diam ^2(\mathcal{Y}) + \epsilon)}\right)
\end{align*}
For $\epsilon<1$ we have
\begin{equation}
\mathbb{P}\left(\left|r_n(x,y_{\alpha})-r^{E}(x,y_{\alpha})\right|>\epsilon\right)\leq 2\exp\left(-Cn\epsilon^2\mathcal{V}_n[x]\right)\quad \mbox{with} \quad C=\frac{\rho_{\min}}{2\diam ^2(\mathcal{Y})(1+\diam ^2(\mathcal{Y}))}
\label{eq:const}
\end{equation}
We deduce from Equation \eqref{eq:all} and Lemma~\ref{lem1}
\begin{equation}
\label{ineq:maxcover}
\mathbb{P}\left(\underset{\alpha=1,\ldots,Q}\max\ \left|r_n\left(x,y_{\alpha}\right)-r^{E}\left(x,y_{\alpha}\right)\right|>\epsilon\right)\leq 2\left(\frac{2\diam(\mathcal{Y})}{\delta}\right)^{\ddim(\mathcal{Y})}\exp \left(-Cn\epsilon^2\mathcal{V}_n[x]\right)
\end{equation}
Let us now bound the term $(iii)$ in \eqref{eq:decomp} as follows
\begin{align}
\label{ineq:diam2}
\left|r^{E}\left(x,y_{\alpha}\right)-r^{E}\left(x,y\right)\right|&=\frac{\mathbb{E}\left[\left(d^2(Y,y_{\alpha})-d^2(Y,y)\right)\textbf{1}\{X\in \pi_n[x]\}\right]}{\mathbb{P}\left(X\in \pi_n[x]\right)}\\
& \leq \frac{\mathbb{E}\left[\left|\left(d(Y,y)-d(Y,y_{\alpha})\right)\left(d(Y,y)+d(Y,y_{\alpha})\right)\right|\textbf{1}\{X\in \pi_n[x]\}\right]}{\mathbb{P}\left(X\in \pi_n[x]\right)} \\
& \leq 2\diam\left(\mathcal{Y}\right)\delta.
\end{align}

We combine inequalities \eqref{ineq:diam1}, \eqref{ineq:maxcover} and \eqref{ineq:diam2} such that with probability, we have:  $1-2\exp\left(\ddim(\mathcal{Y})\log\frac{2\diam(\mathcal{Y})}{\delta}- Cn\epsilon^2\mathcal{V}_n[x]\right)$.

\begin{align}\underset{y\in\mathcal{Y}}\sup\left|r_n(x,y)-r^E(x,y)\right|&\leq \diam(\mathcal{Y})\delta\left(3\mathbb{P}(\mathcal{E}_n)+2\frac{1-\mathbb{P}(\mathcal{E}_n)}{\mathbb{P}(X\in \pi_n[x])}\right) +\epsilon +2\diam(\mathcal{Y})\delta\notag\\ &\leq \diam(\mathcal{Y})\delta\left(3+2\frac{1-\mathbb{P}(\mathcal{E}_n)}{\rho_{\min}\mathcal{V}_n[x]}\right) +\epsilon+2\diam(\mathcal{Y})\delta\notag\\ &=  \diam(\mathcal{Y})\delta\left(5+2\frac{1-\mathbb{P}(\mathcal{E}_n)}{\rho_{\min}\mathcal{V}_n[x]}\right) +\epsilon \notag\\ & \leq \diam(\mathcal{Y})\delta \left(5+8\frac{\Delta_n
^*(\Pi_n)2^{\mathcal{C}(\Pi_n)}\exp -n/128}{\rho_{\min}\mathcal{V}_n[x]}\right)+\epsilon.
\label{ineq:sup3}
\end{align}

Thanks to the assumptions $\frac{\mathcal{C}(\Pi_n)}{n}\rightarrow 0$, $\frac{\log(\Delta_n^*(\Pi_n))}{n}\rightarrow 0$ and $\frac{\log \mathcal{V}_n[x]}{n}\rightarrow 0$ of Theorem~\ref{T1}, the term $\frac{\Delta_n
^*(\Pi_n)2^{\mathcal{C}(\Pi_n)}\exp -n/128}{\rho_{\min}\mathcal{V}_n[x]}$ appearing in the right hand side of the Inequality \eqref{ineq:sup3} converges to zero. Hence, there is a constant $D$ such that $$\frac{\Delta_n
^*(\Pi_n)2^{\mathcal{C}(\Pi_n)}\exp -n/128}{\rho_{\min}\mathcal{V}_n[x]}\leq D$$ for every $n$. Thus we deduce the following inequality that holds with probability $1-2\exp\left(\ddim(\mathcal{Y})\log\frac{2\diam(\mathcal{Y})}{\delta}-Cn\epsilon^2\mathcal{V}_n[x]\right)$

\begin{equation}
\underset{y\in\mathcal{Y}}\sup\left|r_n(x,y)-r^E(x,y)\right|\leq B\delta +\epsilon
\end{equation}
with $B=\diam(\mathcal{Y})(5+8D)$.\\

Let $s>0$ and $\delta=n^{-s}$, for $s$ large enough $B\delta$ is bounded by $\epsilon$. Thus, for $s$ large enough we deduce that
\begin{equation}
\label{ineq:final3}
\mathbb{P}\left(\underset{y\in\mathcal{Y}}\sup\left|r_n(x,y)-r^E(x,y)\right|>2\epsilon\right)\leq 2\exp\left(\ddim(\mathcal{Y})\log\frac{2\diam(\mathcal{Y})}{\delta}-Cn\epsilon^2\mathcal{V}_n[x]\right)
\end{equation}
Under the assumption on $\frac{1}{\mathcal{V}_n[x]}=o\left(\frac{ n}{\log n}\right)$, the probability upper bound on the right hand side in Inequality \eqref{ineq:final3} becomes summable over $n$. We thus conclude the almost sure convergence of $ \underset{y\in\mathcal{Y}}\sup\left|r_n(x,y)-r^{E}(x,y)\right|$ towards zero by the Borel-Cantelli  Lemma.

Finally, we analyze the term $(iv)$ in \eqref{eq:decomp} $ \left|r\left(x,y\right)-r^{E}\left(x,y\right)\right|$
For fixed $x_0\in\mathbb{R}^p$ and $y_0\in\mathcal{Y}$, we have \begin{equation}
r\left(x_0,y_0\right) = \mathbb{E}(d^2(Y,y_0)|X=x_0)=\int_{\mathcal{Y}}d^2(y,y_0)\frac{\rho(x_0,y)}{\rho_X(x_0)}dy
\end{equation}
and \begin{align}
r^{E}\left(x_0,y_0\right) & = \mathbb{E}(d^2(Y,y_0)|X\in \pi_n[x_0]) \notag\\
& = \int_{\mathcal{Y}}d^2(y,y_0)\left(\int_{\pi_n[x_0]}\frac{\rho(x,y)}{\mathbb{P}(X\in \pi_n[x_0])}dx\right)dy\notag\\
&=\int_{\pi_n[x_0]\times\mathcal{Y}}d^2(y,y_0)\rho(x,y)dxdy\times \frac{1}{\mathbb{P}(X\in \pi_n[x_0])}
\end{align}
Moreover, \begin{equation}
\int_{\pi_n[x_0]}\frac{\rho(x,y)}{\mathbb{P}(X\in \pi_n[x_0])}dx=\frac{\int_{\mathbb{R}} \textbf{1}\left\{x\in \pi_n[x_0]\right\}\rho(x,y)dx}{\int_{\mathbb{R}} \textbf{1}\left\{x\in \pi_n[x_0]\right\}\rho_X(x)dx}
\end{equation}
Since $\rho$ is uniformly continuous, for every $(x_0,y)\in\mathbb{R}^p\times\mathcal{Y}$, $\forall \epsilon>0,\exists\delta^1_{\epsilon}>0$ such that $||x_0-x||\leq \delta^1_{\epsilon} \Rightarrow |\rho(x_0,y)-\rho(x,y)|\leq \epsilon$. Thus, there exists $\delta^1_{\epsilon}>0$ such that 
\begin{align}
\label{ineq:vol}&\left|\int_{\mathbb{R}}\textbf{1}\left\{x\in \pi_n[x_0]\right\}  (\rho(x,y)-\rho(x_0,y))dx \right| \notag \\ &\leq \int_{\pi_n[x_0]}|\rho(x,y)-\rho(x_0,y)|dx \notag\\ &=\int_{B(x_0,\delta^1_{\epsilon})\cap \pi_n[x_0]}|\rho(x,y)-\rho(x_0,y)|dx + \int_{\pi_n[x_0]/B(x_0,\delta^1_{\epsilon})}|\rho(x,y)-\rho(x_0,y)|dx\notag \\ &\leq \epsilon\Vol\left(\pi_n[x_0]\cap B(x_0,\delta^1_{\epsilon})\right) + 2||\rho(.,y)||_{\infty}\Vol\left(\pi_n[x_0]\backslash B(x_0,\delta^1_{\epsilon})\right)\notag\\ &\leq \epsilon\Vol\left(\pi_n[x_0]\right) + 2||\rho(.,y)||_{\infty}\Vol\left(\pi_n[x_0]\backslash B(x_0,\delta^1_{\epsilon})\right)
\end{align}
Using the same argument of continuity on the density $\rho_X$, for all $\epsilon$, there is $\delta_{\epsilon}^2$ such that 
\begin{equation}
\label{ineq:volrho2}
\int_{\mathbb{R}}\textbf{1}\left\{x\in \pi_n[x_0]\right\}\left(\rho_X(x)-\rho_X(x_0)\right)dx\leq \epsilon\Vol(\pi_n[x_0])+2||\rho_X||_{\infty}\Vol\left(\pi_n[x_0]\backslash B(x_0,\delta^2_{\epsilon})\right)
\end{equation}
We define $\delta_{\epsilon}=\min(\delta_{\epsilon}^1,\delta_{\epsilon}^2)$. 
We will apply the dominated convergence theorem to conclude.
To this end, we remark that for every sequence of functions $(f_n)_n$, $(g_n)_n$ and for every functions $f$ and $g$ we have 
\begin{align*}
 \left|\frac{f_n}{g_n} - \frac{f}{g}\right|& = \left|\frac{f_n}{g_n} - \frac{f}{g_n} + \frac{f}{g_n} - \frac{f}{g}\right|\\ &= \left|\frac{f_n-f}{g_n}  - f\frac{g-g_n}{gg_n}\right|\\ &\leq \frac{\left|f_n-f\right|}{g_n} + f\frac{\left|g-g_n\right|}{gg_n}
\end{align*}
We take \begin{equation*}
f_n(x_0,y)=\int_{\pi_n[x_0]}\rho(x,y)dx;\qquad f(x_0,y)=\rho(x_0,y);\\
\end{equation*}
\begin{equation*}
g_n(x_0)=\int_{\pi_n[x_0]}\rho_X(x)dx;\qquad g(x_0)=\rho_X(x_0).
\end{equation*}
We deduce the following upper bound
\begin{align}\label{ineq:f}
\frac{\left|f_n-f\right|}{g_n}&= \frac{\left|\int_{\pi_n[x_0]}\rho(x,y)dx-\rho(x_0,y)\right|}{\int_{\pi_n[x_0]}\rho_X(x)dx} \notag\\ &\leq \frac{\left|\int_{\pi_n[x_0]}\rho(x,y)dx-\rho(x_0,y)\right|}{\rho_{\min}\Vol(\pi_n[x_0])} \notag\\&\leq \frac{\epsilon\Vol(\pi_n[x_0])+2||\rho(.,y)||_{\infty}\Vol\left(\pi_n[x_0]\backslash B(x_0,\delta_{\epsilon})\right)}{\rho_{\min}\Vol(\pi_n[x_0])}\notag\quad\mbox{using}\ \eqref{ineq:vol}\\ &= \frac{\epsilon}{\rho_{\min}}+\frac{2||\rho(.,y)||_{\infty}\Vol\left(\pi_n[x_0]\backslash B(x_0,\delta_{\epsilon})\right)}{\rho_{\min}\Vol(\pi_n[x_0])}
\end{align}
with the same arguments we also get using~\eqref{ineq:volrho2}
\begin{equation}\label{ineq:g}
\frac{\left|g_n-g\right|}{g_n}=\frac{\left|\int_{\pi_n[x_0]}\rho_X(x)dx-\rho_X(x_0)\right|}{\int_{\pi_n[x_0]}\rho_X(x)dx}\leq \frac{\epsilon}{\rho_{\min}}+\frac{2||\rho_X||_{\infty}\Vol\left(\pi_n[x_0]\backslash B(x_0,\delta_{\epsilon})\right)}{\rho_{\min}\Vol(\pi_n[x_0])}
\end{equation}
From the assumptions of Theorem~\ref{T1}, we have that $\diam(\pi_n[x_0])$ converges towards zero almost surely. Hence, with probability 1, for every $\delta_{\epsilon}$, there is $N_{\epsilon}>0$ such that for every $n\geq N_{\epsilon}$, $\diam(\pi_n[x_0])\leq \delta_{\epsilon}/2$. Thus, for every $n\geq N_{\epsilon}$, $\Vol ( \pi_n(x_0)\backslash B (x_0,\delta_{\epsilon}))=0$ almost surely. Then from \eqref{ineq:f} and \eqref{ineq:g} we deduce that for every $n\geq N_{\epsilon}$ the following inequalities hold almost surely

\begin{equation}\label{ineq:fg}
\frac{\left|f_n-f\right|}{g_n}\leq \frac{\epsilon}{\rho_{\min}} \quad \mbox{and}\quad \frac{\left|g_n-g\right|}{g_n}\leq \frac{\epsilon}{\rho_{\min}}
\end{equation}
Finally, we deduce from \eqref{ineq:fg}

 \begin{equation}
\left|\frac{f_n}{g_n}-\frac{f}{g}\right|\leq \frac{\left|f_n-f\right|}{g_n} + f\frac{\left|g-g_n\right|}{gg_n}\leq \frac{(f+g)\epsilon}{g\rho_{\min}}\qquad \mbox{a.s}
\end{equation}
Moreover \begin{equation}
\left|\frac{f_n}{g_n}\right|\leq \frac{||\rho||_{\infty}\Vol(\pi_n[x])}{\rho_{\min}\Vol(\pi_n[x])}=\frac{||\rho||_{\infty}}{\rho_{\min}}<\infty\quad \mbox{from \textbf{P1}}
\end{equation}
Using the dominated convergence theorem we thus get
\begin{equation}
\lim_{n\rightarrow+\infty}r^E(x_0,y_0)= \int_{\mathcal{Y}}d^2(\omega,y_0)\frac{\rho(x_0,\omega)}{\rho(x_0)}d\omega=r\left(x_0,y_0\right)\qquad \mbox{with probability 1}.
\end{equation}
Finally, we demonstrate the weak consistency given by \eqref{eq:convmoy}. The proof uses the arguments from \cite{hein2009robust}. Under the  assumptions of Theorem \ref{T1},    for every $x\in\mathbb{R}^p$, one has that, $\lim\limits_{n\rightarrow\infty}r\left(x,T_n(x)\right)=r\left(x,\phi^*(x)\right)$ almost surely. Now, remark that 
$$
R(T_n)-R(\phi^*)\leq\mathbb{E}\left(\left|r\left(X,T_n(X)\right)-r\left(X,\phi^*(X)\right)\right|\right).
$$ 
As $\diam\left(\mathcal{Y}\right)<\infty$, we have that $\mathbb{E}\left(r\left(X,T_n(X)\right)\right)<+\infty$ and $\mathbb{E}\left(r\left(X,\phi^*(X)\right)\right)<+\infty$. Therefore, an extension of the dominated convergence theorem given  in \cite{glick1974consistency} allows to conclude.

\section{Complements about the first simulation scenario}
\label{Complement3}
\subsection{Three temporal behavior functions scheme}

In this case the simulation model of the input curves is given as follows 

\begin{equation}
X_i^{(j)} (t) = \begin{cases} \beta_i \left( f_{j,1} (t) \textbf{1}_{\{G^j_i=1\}} +
f_{j,2} (t) \textbf{1}_{\{G^j_i=2\}}  + f_{j,3} (t) \textbf{1}_{\{G^j_i=3\}}\right) + W_i^1 (t) \quad \mbox{if }j\in\{1,2\}\\
\beta_i' \left( f_{j,1} (t) \textbf{1}_{\{G'^j_i=0\}} +
f_{j,2} (t) \textbf{1}_{\{G'^j_i=2\}} + f_{j,3} (t) \textbf{1}_{\{G'^j_i=3\}}\right) + W_i^1 (t)\quad \mbox{if }j\in\{3,4,5,6\}\end{cases}
\label{simX3}
\end{equation}
where all the parameters $p$, $n$, $\beta_i$, $W_i^1(t)$, $f_{j,1}$ and $f_{j,2}$ remain identical to the previous scheme;  $G^j_i$ and $G'^j_i\sim \mathcal{U}\left(\left\{1,2,3\right\}\right)$, $f_{j,3}$ are
defined as follows:
\begin{equation*}
\begin{cases}
			f_{1,3}(t) = 0.5t + 0.1 \sin(6t)  \\
			f_{2,3}(t) = 0.3 - 0.7 (t-0.45)^2 \\
			f_{3,3}(t) = 2 (t - 0.5)^2 - 0.3 t \\
			f_{4,3}(t) = 0.2 - 0.3t + 0.1 \cos(8t) \\
			f_{5,3}(t)=f_{1,1}(t) \\
			f_{6,3}(t)=f_{1,2}(t) \\
\end{cases}
\label{eq:tempoBehav3}
\end{equation*}
Similarly, the output variable $Y$ is simulated according to the different pairs of temporal behaviors for the the first input variables, hence $(G_i^1,G_i^2)$ is used
to determine a trajectory for the output variable:
\begin{equation}
Y_i(t) = \beta_i \sum_{j=1}^3 \sum_{k=1}^3 g_{j,k}(t) \textbf{1}_{\{G^1_i=j\}} 
\textbf{1}_{\{G_i^2=k\}} + W_i^2(t)
\label{simY3}
\end{equation}
where $g_{j,k}$ being identical to the previous scheme for all $j,k \in \{1,2\}$ and 
\begin{equation}
\left\{
\begin{array}{l@{\hskip 5ex}l}
            g_{1,3}(t) =   0.2 - 0.3t +0.1\cos (8t)  \\
            g_{2,3}(t) =   (t-0.42)^2                \\
            g_{3,3}(t) =   0.15 + 0.7t\sin (3 t)     \\
            g_{3,2}(t) =   0.5t^2 - 0.2\sin (5t)     \\
            g_{3,1}(t) =   0.6\log (t+1) +0.3\sin (5t)\\
\end{array}
\right.
\label{eq:behavY3}
\end{equation}

\begin{figure*}[!ht]
\begin{center}
\includegraphics[width=\textwidth]{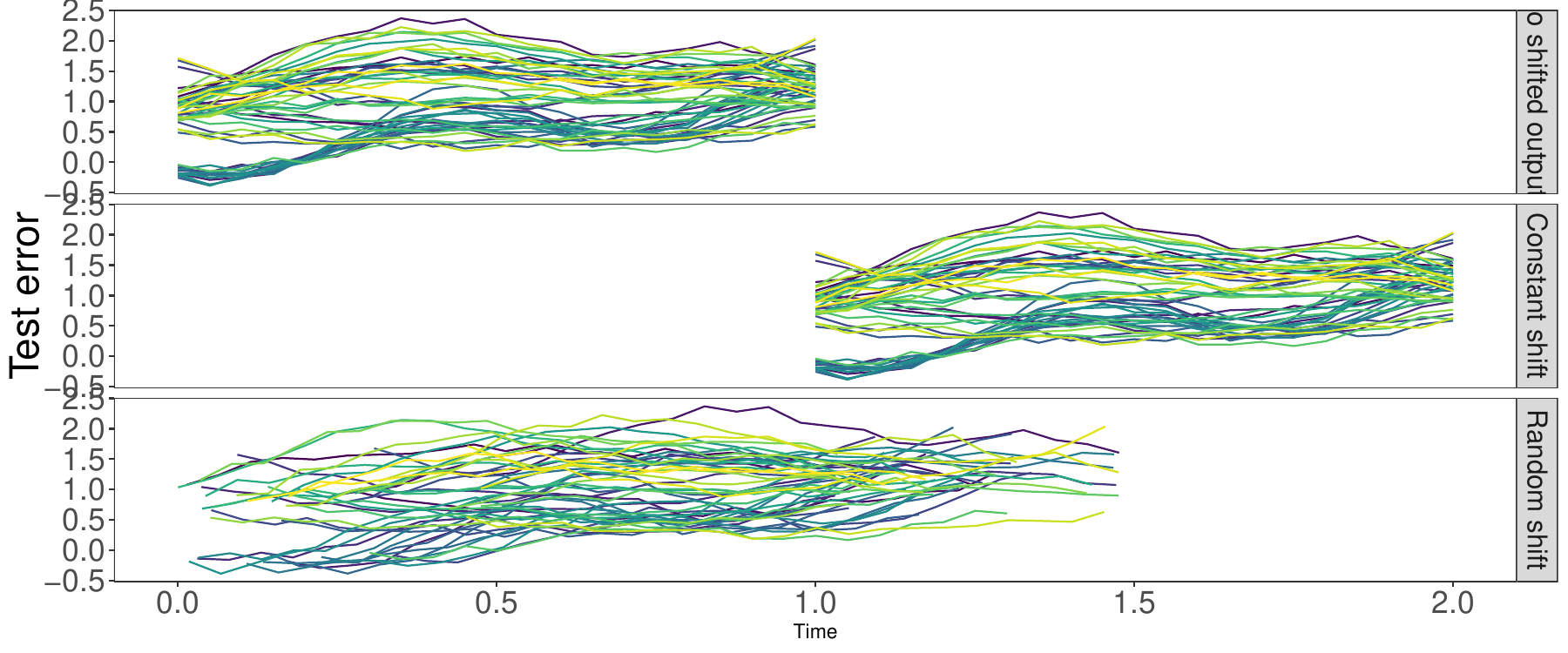}
\caption{Dynamics of the output variable curves simulated according to the model~\eqref{simY} in the standard case (i.e. without time shift), with a constant time shift equal to 1; with a uniform time shift $\mathcal{U}([0,1])$.}
\label{compY}
\end{center}
\end{figure*}

\section{Predict images with curves, a toy example}
\label{app:AN2}
\subsection{Simulation scheme}

The purpose of this scenario is to illustrate the ability of the Fréchet RF method to predict images from input curves. We simulate a dataset of $n=500$ observations, the input curve variables are simulated according to the model~\eqref{simX} of the first scenario.  As in the second scenario, the output images are taken from the MNIST dataset \citep{lecun-mnisthandwrittendigit-2010}. For any $k=1,\ldots,8$ and for any $i=1,\ldots,n$ we note $\mathcal{M}_i^k$ the random draw of the handwritten $k$ digit in the MNIST dataset for the $i$th observation. Let the pair $(G_i^1,G^2_i)$ used to attribute their shape to the curves of the first two input variables for the $i$th observation and $\beta_i$ the expansion/contraction parameter of these same curves, then the output images are drawn according to the combinations summarized in the Table~\ref{combo}.

\begin{table}[!ht]
\centering
\begin{tabular}{|c|c|c|c|c|c|}
\hline
& $G^1_i=1,\ G_i^2=1$ & $G^1_i=2,\ G_i^2=1$ & $G^1_i=1,\ G_i^2=2$ & $G^1_i=2,\ G_i^2=2$  \\
\hline
$\beta_i >1$ & $\mathcal{M}_i^1$ & $\mathcal{M}_i^3$ & $\mathcal{M}_i^5$ & $\mathcal{M}_i^7$\\
\hline
$\beta_i\leq 1$ &  $\mathcal{M}_i^2$ &  $\mathcal{M}_i^4$ & $\mathcal{M}_i^6$ & $\mathcal{M}_i^8$    \\
\hline
\end{tabular}
\caption{Random draws in the MNIST dataset of the output images from the realizations $G^1_i,\ G^2_i$ and $\beta_i$ used to simulate the input curves.}
\label{combo}
\end{table}
As in the first scenario, the output images depend only on the first two input variables, the link between the images and the curves is entirely contained in the pairs $(G_i^1,G_i^2)$ as well as in the $\beta_i$. This means that the handwritten number images then depend both on the shape of the curves of the first two input variables and their amplitude.

\subsection{Results}

\begin{figure}[!ht]
\centering
\includegraphics{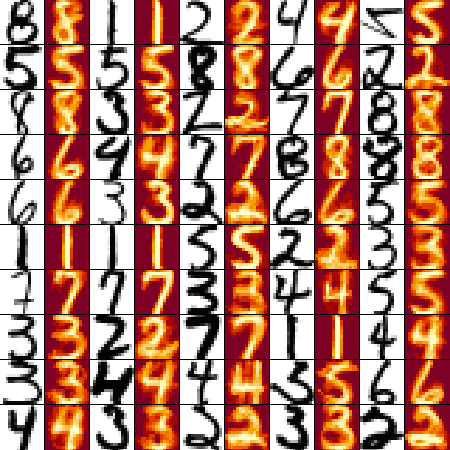}
\caption{True output images and OOB predictions. In black and white (grayscale), 50 output images from the dataset of $n=500$ observations simulated according to the third scenario are displayed. The redscale image to the right of each grayscale image is the OOB prediction given by the trained Fréchet RF. }
\label{fig:PredIm}
\end{figure}

The Fréchet distance is used on the curve spaces, i.e. on the 6 input variables. The distance used on the output space is the standard Euclidean distance. A Fréchet RF is constructed with $q=500$ trees (justified by the fact that the OOB error of the Fréchet RF becomes stable as long as 350 trees compose the forest). Similarly, the \texttt{mtry} parameter is set to 5. 
As shown in Figure~\ref{fig:PredIm}, OOB predictions of output images always give the correct written digit. However, ghosting can be seen on some digit predictions. This is due to the simulation scheme itself. The input curves only give information about the written number, and do not provide any information about its individual characteristics such as the width of the number, its height, the presence or not of a loop (for writing a 2 for example). More precisely, there is within the same group of numbers (for example the set of numbers 4 drawn) a variability in the written numbers that is not explained by the input curves. By introducing variables that provide information on the fine characteristics of each written number (such as its height, width, etc.) we would get even more accurate predictions. Moreover, it is noticeable that this phenomenon of ghosting is not present for numbers that have a very low variability in their writing such as the number 1. In order to highlight this point a Fréchet RF is constructed on the same simulated dataset and the images are replaced by factors indicating what the written number is. When the outputs are images, the percentage of explained variance is 20\%, which was expected since there is a large variability between the same numbers that is not explained by the input curves. When the outputs are factors expressing the written numbers, the percentage of explained variance is 98\%. It is therefore clear that the link between the output images and the input curves relates only to the digits and not to its individual characteristics. So even if the explained variance percentage is only 20\% for the images, the Fréchet RF (almost) always predicts the right digit. 

\section{Air quality prediction results}
\label{ANairqual}

\begin{figure}[!ht]
\centering
\includegraphics{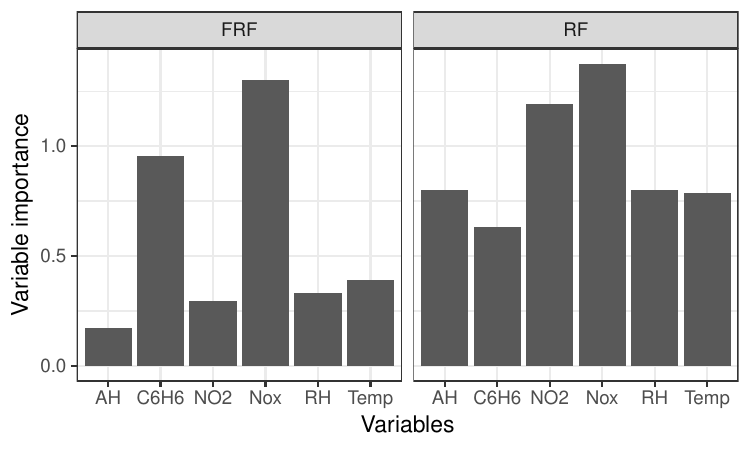}
\caption{Variable importance scores for FRF and RF predictors applied on the \texttt{airquality} dataset.}
\label{fig:varImpComp}
\end{figure}

In Figure~\ref{fig:varImpComp} we plotted the variable importance scores associated to FRF and RF predictors used for the analysis of the \texttt{airquality} dataset.

\vskip 0.2in
\bibliographystyle{apalike}
\bibliography{frec}

\end{document}